\documentclass[10pt,journal,compsoc]{IEEEtran}
\ifCLASSOPTIONcompsoc
\usepackage[nocompress]{cite}
\else
\usepackage{cite}
\fi

\usepackage{graphicx,subfig,multirow,array}

\usepackage{amsfonts,amssymb}
\usepackage[cmex10]{amsmath}
\usepackage{algorithmic}
\usepackage[ruled]{algorithm2e}	
\usepackage{epstopdf}
\usepackage{hyperref}

\usepackage{verbatim}
\usepackage[usenames,dvipsnames]{color}
\usepackage{soul}

\DeclareMathOperator*{\argmin}{arg\,min}

\begin{document}
\title{Robust Video Content Alignment and Compensation for Clear Vision Through the Rain}

\author{Jie~Chen$^\S$,~\IEEEmembership{Member,~IEEE,}
	Cheen-Hau~Tan$^\S$,~\IEEEmembership{Member,~IEEE,}
	Junhui~Hou,~\IEEEmembership{Member,~IEEE,}	
	Lap-Pui~Chau,~\IEEEmembership{Fellow,~IEEE,}
	and~He~Li,
	\IEEEcompsocitemizethanks{
		\IEEEcompsocthanksitem $\S$: J. Chen and C.-H. Tan contributed equally to this work.
		\IEEEcompsocthanksitem J. Chen, C.-H. Tan, and L.-P. Chau are with the School of Electrical and Electronic Engineering, Nanyang Technological University, Singapore. E-mail: \{chen.jie, cheenhau, elpchau\}@ntu.edu.sg.
		\IEEEcompsocthanksitem J. Hou is with  Department of Computer Science, City University of Hong Kong. E-mail: jh.hou@cityu.edu.hk.
		\IEEEcompsocthanksitem H. Li is with Singapore Technologies Dynamics Pte Ltd. E-mail: lihe@stengg.com.
		\IEEEcompsocthanksitem The research was partially supported by the ST Engineering-NTU Corporate Lab through the NRF corporate lab@university scheme.\protect\\
	}% <-this % stops an unwanted space
}
	
% \markboth{Journal of \LaTeX\ Class Files,~Vol.~14, No.~8, August~2015}%
%\markboth{Manuscript Submitted to IEEE Transactions on Pattern Analysis and Machine Intelligence}%
% {Shell \MakeLowercase{\textit{et al.}}: Bare Demo of IEEEtran.cls for Computer Society Journals}
%{Robust}

\IEEEtitleabstractindextext{%
\begin{abstract}
Outdoor vision-based systems suffer from atmospheric turbulences, and rain is one of the worst factors for vision degradation. Current rain removal methods show limitations either for complex dynamic scenes, or under torrential rain with opaque occlusions. 
We propose a novel derain framework which applies superpixel (SP) segmentation to decompose the scene into depth consistent units.
Alignment of scene contents are done at the SP level, which proves to be robust against rain occlusion interferences and fast camera motion.
Two alignment output tensors, i.e., optimal temporal match tensor and sorted spatial-temporal match tensor, provide informative clues for the location of rain streaks and the occluded background contents.
Different classical and novel methods such as Robust Principle Component Analysis and Convolutional Neural Networks are applied and compared for their respective advantages in efficiently exploiting the rich spatial-temporal features provided by the two tensors. 
Extensive evaluations show that advantage of up to 5\textit{dB} is achieved on the scene restoration PSNR over state-of-the-art methods, and the advantage is especially obvious with highly complex and dynamic scenes. Visual evaluations show that the proposed framework is not only able to suppress heavy and opaque occluding rain streaks, but also large semi-transparent regional fluctuations and distortions.

\end{abstract}
	
\begin{IEEEkeywords}
superpixel segmentation, rain removal, dynamic content, spatial-temporal filtering, robust principle component analysis, convolutional neural networks.
\end{IEEEkeywords}}

\maketitle
	
\IEEEdisplaynontitleabstractindextext

% \fi
\IEEEpeerreviewmaketitle

% The fist section name appear above text in both columns in the first page for all computer society papers including PAMI.
%\IEEEraisesectionheading{\section{Introduction}\label{sec_introduction}}
\section{Introduction}\label{sec_introduction}
%\IEEEPARstart{M}{odern} intelligent systems rely more and more on information from cameras as input, the quality of which directly affects the robustness of the whole system. For outdoor vision modules, atmospheric turbulences could cause serious degradation on the input image/video quality. 
%Depending on the size and type of atmospheric particles, weather conditions can be generally classified into steady or dynamic systems \cite{Tripathi2014Removal}. For steady systems such as fog and haze, atmospheric particles (range between 1-10 $\mu m$) suspend in the air and influence the vision systems via accumulated scattering of the scene radiation, \textcolor{Red}{which results in serious degradation of vision contrast.} Statistical modeling of the scene depth helps to estimate the weather free details \cite{he2009single, fattal2014dehazing, berman2016non}. For dynamic weather systems such as rain and snow, involved particles are usually much larger (0.1-10 $mm$) with fast \textcolor{Red}{motion} (caused by gravity, wind etc.). They could introduce false content motion, and seriously degrade contrast and visibility, and obscure important features for other computer vision algorithms. Rain removal is therefore a vital procedure for the weather-robustness of most outdoor vision based systems.

%\IEEEPARstart{M}{odern} intelligent systems rely more and more on visual information, the quality of which directly affects the robustness of the whole system. 

\IEEEPARstart{A}{s} modern intelligent systems rely more and more on visual information, the robustness of the system becomes dependent on the quality of visual input.
Outdoor vision modules, in particular, are susceptible to visual degradations caused by atmospheric turbulences.
Turbulent weather conditions can be generally classified into \textit{steady} or \textit{dynamic} systems \cite{narasimhan2002vision, Garg2007}. 
For steady systems, such as fog and haze, small (1-10 $\mu m$) atmospheric particles suspended in the air induce accumulated scattering of scene radiance; hence, they primarily degrade visual contrast and visibility.
For dynamic systems, such as rain and snow, the atmospheric particles are much larger (0.1-10 $mm$) and relatively fast-moving.
As such, the motions of individual or aggregate particles become visible and cause complex visual distortions.
These distortions could unintentionally introduce motion content or obscure visual features, which negatively impacts computer vision algorithms.
In this paper, we propose a rain removal framework that handles both the dynamic and steady state visual degradations of rain.

% Introduce physical properties of rain here
To devise a rain removal algorithm, we should first understand the visual effects caused by rain.
%The raindrops fall at a constant terminal velocity when it descends to a low altitude due to air resistance \cite{Gunn1949terminal}. Although the terminal velocity varies with respect to raindrop sizes, their speed are all rather fast when captured by a normal frame rate camera, i.e., significant location differences for the same raindrop will be observed between adjacent frames.
Rain appearances can be significantly different depending on scene distance ranges and imaging configurations \cite{Garg2005When}. 
Based on their visual effects, we classify rain into three types.
As shown in Fig.~\ref{fig_rainTypes}, these are:
\begin{itemize}
	\item \textit{\textbf{Occluding rain:}}
	these raindrops fall within the depth of field of a camera. They are in focus, and fall at high speed. Hence, they appear as slender streaks under normal camera exposure settings (shutter speed slower than 1/200 second). In general, these streaks are somewhat opaque and occlude the background scene.
	
	\item \textit{\textbf{Veiling rain:}}
	raindrops that fall nearer than the current depth of field cover a larger field of view and are out of focus. 
	They might not take distinct shapes and resemble semi-transparent veils, causing regional intensity fluctuations and distortions.
	
	\item \textit{\textbf{Accumulated rain:}}
	raindrops distributed further than the current depth of field are out of focus, cover a tiny field of view, and are distributed over large distances. 
	Their visual effects are accumulated over many samples and are stable over time, producing visual degradations similar to that of fog and haze.
\end{itemize}

%%%%%%%%%%%%%%%%%%%%%%%Figure%%%%%%%%%%%%%%%%%%%%%%%
\begin{figure}[!t]
	\centering
	\includegraphics[width=0.88\linewidth]{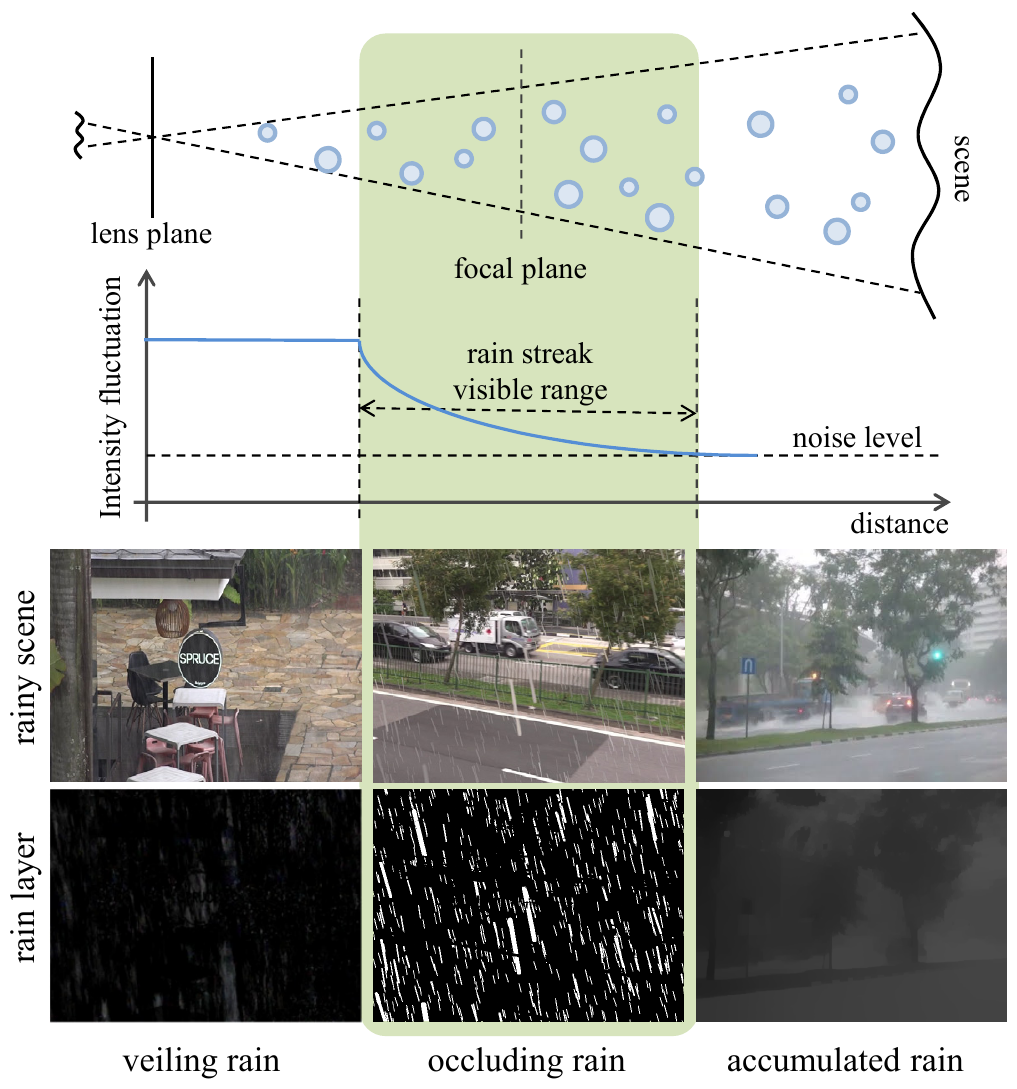}
	\caption{Different photometric appearances of raindrops at different distance range and camera configurations.}
\end{figure}\label{fig_rainTypes}
%%%%%%%%%%%%%%%%%%%%%%%Figure%%%%%%%%%%%%%%%%%%%%%%%

Based on the types of input, we can categorize rain removal methods into image-based methods and video-based methods.
Image-based methods rely only on information from the processed frame while video-based methods have access to other frames in the sequence.
Most video-based methods rely primarily on temporal information since they are very discriminative for rain detection -- both occluding rain and veiling rain produce intensity fluctuations over time; other clues like rain structure might not consistently hold under all rainy conditions, e.g., under veiling rain.

However, temporal information are reliable only if, for each pixel, the corresponding scene content is consistent over time.
%However, temporal information are reliable only if the analyzed scene is consistent over consecutive frames.
This constraint could be violated due to two factors -- motion of the camera and presence of moving objects.
Previous works tackle these two issues separately.
Camera motion-induced scene content shifts were reversed using global frame alignment \cite{Tripathi2012, tan2014dynamic, ren2017video}.
However, the granularity of global alignment is too large when scene depth range is large; 
to be aligned accurately, scene content of different depths should be independently aligned.
Thus, by using global alignment, parts of the scene could be poorly aligned.

The presence of moving objects disrupts temporal consistency of scene content, and could thus cause scene elements to be misclassified as rain.
Earlier works attempt to identify the misclassified scene pixels by testing various properties of candidate rain pixels, such as photometric properties \cite{Garg2004}, chroma changes \cite{zhang2006}, geometric properties \cite{bossu2011rain} or local statistics \cite{Tripathi2012}.
This approach, however, does not adequately remove rain that overlap moving objects.
Later works first segment out motion regions or foreground objects before applying rain removal on these regions separately \cite{chen2014rain,ren2017video,wei2017should}.
However, these methods rely on accurate segmentation of foreground regions.

We recognize that both camera motion and moving objects cause scene content to be misaligned and thus corrupt temporal information.
%disrupt their consistency
Hence, in this paper, we propose a novel and elegant framework that simultaneously solves both issues by aligning scene content to compensate for their misalignment -- superpixel-based content alignment and compensation (SPAC) for rain removal.
First, the target video frame is segmented into superpixels (SP)s; each SP corresponds to a depth consistent patch of scene content.
Then, for each target SP, we form tensors by stacking patches similar to the target SP.
This step aligns content corresponding to both scene background and moving objects without prior segmentation of the moving object.
Scene content is also much better aligned at a SP-level granularity.
Finally, a de-rained SP patch is extracted from the tensors.
Within this framework, we experimented on several ways to utilize these tensors for rain removal -- taking the average values of the tensor (\textit{SPAC-Avg} and \textit{SPAC-$\mathcal{F}_1$}) , applying robust principle analysis (\textit{SPAC-RPCA}), and processing it using a Convolutional Neural Network (\textit{SPAC-CNN}).

\begin{figure*}[!t]
	\centerline{\subfloat{\includegraphics[width=0.84\linewidth]{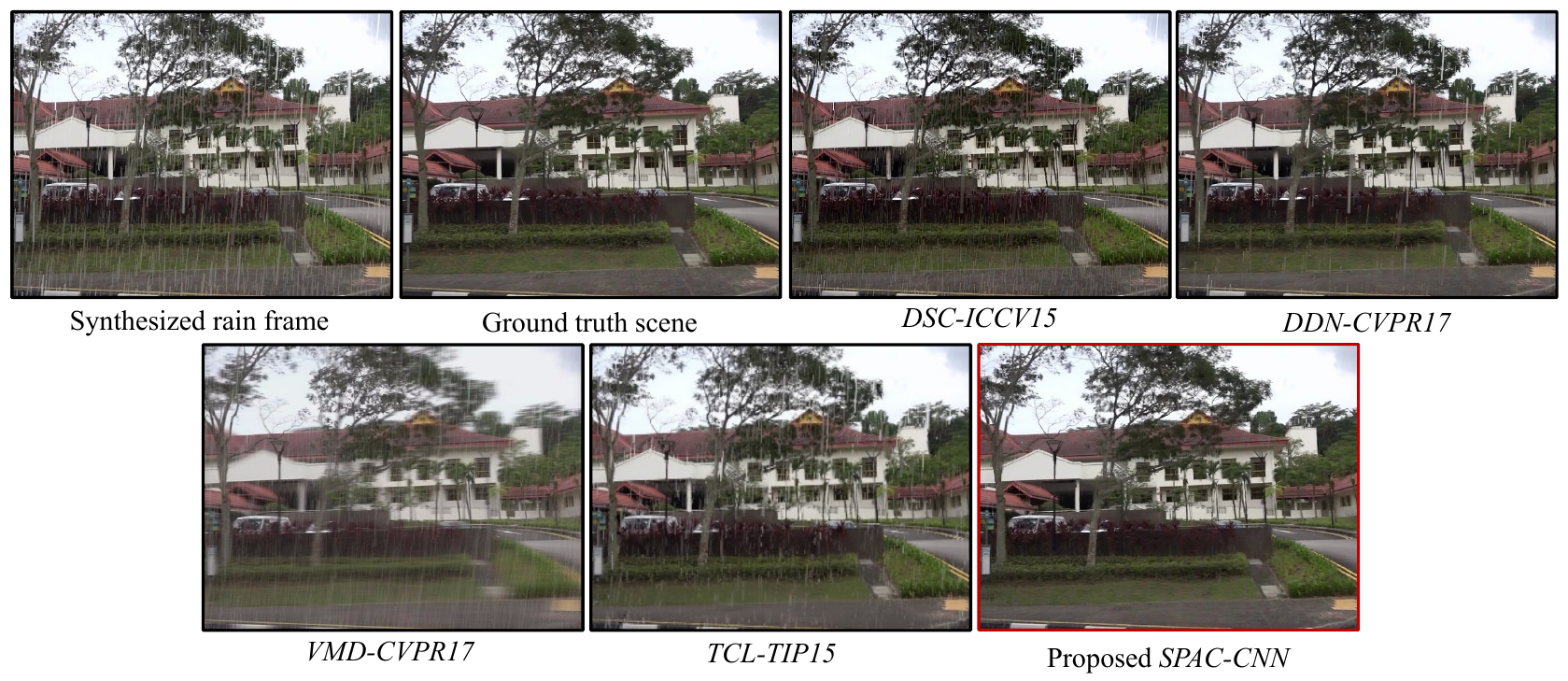}}}
	\caption{Comparison of derain outputs by different algorithms for a challenging video sequence with fast camera motion and heavy rain fall. Image-based derain methods, i.e., discriminative sparse coding (\textit{DSC}) \cite{luo2015removing} and deep detail network (\textit{DDN}) \cite{fu2017removing} fail to remove large and/or opaque rain streaks. Video-based derain methods, i.e., via matrix decomposition (\textit{VMD}) \cite{ren2017video}, and via temporal correlation and low-rank completion (\textit{TCL}) \cite{kim2015video}, create serious blur due to fast camera motion, and also show significant amount of remnant rain streaks. Our proposed \textit{SPAC-CNN} can cleanly remove the rain streaks and preserve scene content truthfully.}
	\label{fig_teaserComp}
\end{figure*}

Extensive evaluations show that up to 5\textit{dB} reconstruction PSNR advantage is achieved over state-of-the-art methods, and the advantage is especially obvious in highly dynamic scenes taken from a fast moving camera. Visual evaluations show that the proposed framework is not only able to remove heavy and opaque occluding rain streaks, but also suppress large semi-transparent regional fluctuations and distortions. 
Fig.~\ref{fig_teaserComp} illustrates the advantage of our proposed algorithm over existing methods in a challenging video sequence.
The contributions of this work can be generalized as follows. \vspace{-0.2cm}

\begin{itemize}
	%	\item We propose a novel spatial-temporal content alignment algorithm at SP level, which can handle fast camera motion and dynamic scene contents in one framework. This mechanism greatly outperforms existing scene motion analysis methods that models background and foreground motion separately.
	%\item The strong local properties of SPs can robustly counter heavy rain interferences, and facilitate much more accurate alignment. Owing to such robust alignment, accurate temporal correspondence could be established for rain occlusions such that heavily occluded backgrounds could be truthfully restored. This greatly outperforms image-based derain methods in which recovery of large and opaque rain occlusions remain the biggest challenge.
	%\item We propose a set of very efficient spatial-temporal features for the compensation of high frequency details lost during the deraining process. An efficient CNN network is designed, and a synthetic rain video dataset is created for training the CNN.
	% The data set contains 150,000 examples with embedded temporal feature tensors for efficient training of the CNN to extract useful high frequency content details to restore the scene fidelity after rain removal.
	\item We propose to apply spatial-temporal alignment on scene content for video-based rain removal. This approach, ensures content consistency over consecutive frames whether subjected to camera motion or presence of moving objects. Hence, segmentation of moving objects for separate treatment is not required.
	\item We propose to use Superpixels as the basic unit for scene content alignment. Alignment of small blocks of pixels is not robust to structured noise, especially large rain streaks. Large blocks of pixels, however, are oblivious to object boundaries, and might encompass disjointed content located at different depths. By using SPs, we can form sufficiently large clusters of pixels that also respect object boundaries.
	\item We tested a set of different rain removal methods on aligned scene content. In particular, we propose a neural network-based (\textit{SPAC-CNN}) method where a set of efficient spatial-temporal features is used to help restore high frequency details lost during the de-raining process.
	\item As far as we know, this is the first work to categorize testing rain data based on camera speed for separate evaluations. A rain video dataset was created with a car-mount camera with speed range between 0 to 30 \textit{km/h} over different dynamic scenes. The data set was used both for model training and evaluations.
\end{itemize}

%%%%%%%%%%%%%%%%%%%%%%%Figure%%%%%%%%%%%%%%%%%%%%%%%
%%%%%%%%%%%%%%%%%%%%%%%Figure%%%%%%%%%%%%%%%%%%%%%%%

A shorter version of this work appeared in \cite{chen2018robust}. This journal version presents more explanations and technical details, and includes additional experiments to verify the efficiency of using superpixels over blocks. We have also included additional rain removal methods, \textit{SPAC-$F_1$} and \textit{SPAC-RPCA}, under our superpixel-based content alignment framework. Comparisons to additional competing video-based methods have also been added. A patent for an implementation based on \textit{SPAC-$F_1$} \cite{chen2017patent} is being filed in Singapore.

The remainder of the paper is organized as follows.
In Section \ref{sec_related}, we will discuss the current rain removal methods.
Then, in Section \ref{sec_proposedModel}, we introduce our proposed framework and methods in detail.
In Section \ref{sec_experiments}, we show comprehensive evaluations of our methods and competing methods, and discuss the results.
Finally, Section \ref{sec_conclusion} concludes the paper.

\section{Related Work}\label{sec_related}
Rain can exhibit dynamic effects and steady state effects.
Steady state rain, which we call accumulated rain, can be removed using haze and fog removal methods \cite{he2009single, fattal2014dehazing, berman2016non}.
For dynamic rain, rain removal methods are usually categorized into video-based and image-based methods.
Alternatively, since rain appearance is a function of camera parameters, it is possible to reduce the visibility of rain by adjusting these parameters \cite{Garg2005When}.
Another related research topic is adherent rain removal \cite{you2016adherent, eigen2013restoring}, which deals with distortions caused by raindrops that adhere on a transparent surface. 

\subsection{Image-based methods}
Rain removal based on a single image is usually modeled as a layer separation problem, where the input image is a composition of a background layer and rain layer.
In order to separate both layers, priors intrinsic to background and rain respectively are required. 
%For example, rain streaks are assumed to be of similar orientation throughout the image, while background edges are not.
Kang et al. \cite{kang2012automatic} proposed to learn a dictionary from the high frequency layer, which contains both background edges and rain streaks, of the input image.
By assuming that rain streaks are of similar orientation throughout the image while background edges are not,
atoms corresponding to rain and background are separated based on their Histogram of Gradients.
The high frequency layer of the output is obtained by applying sparse coding using the background dictionary.
%This approach does not adequately preserve background edges with similar orientations to rain.
Further enhancements to this framework include improving the separation of rain and background atoms \cite{huang2014self, sun2014exploiting}, and considering additional priors such as blurriness and saturation of rain \cite{chen2014visual}.

Chen and Hsu \cite{chen2013generalized} derived an optimization function where that the rain layer is assumed to be low ranked while the background layer varies smoothly.
Luo et al. \cite{luo2015removing} relied on discriminative sparse coding, where their rain dictionary is trained from simulated rain.
Li et al. \cite{li2016rain, li2017single} modeled rain and background layers using Gaussian mixture models (GMM). 
Natural images are used to train the background GMM while pure rain streak regions in the input, which do not contain background features, are used for training the rain GMM.
Instead of applying layer separation, Kim et al.\cite{kim2013single-image} identified rain pixels using local statistical and geometric properties, such as orientation, shape and brightness, and then recovered their values using non-local means filtering.

Convolutional Neural Networks have been very effective in both high-level vision tasks \cite{krizhevsky2012imagenet} and low-level vision applications for capturing signal characteristics \cite{Kim2016accurate,zhang2017beyong}.
Different network structures and features were explored for rain removal. 
Fu et al. proposed the deep detail network \cite{fu2017removing} which uses a priori image domain knowledge to focus on learning high frequency details of the rain and differentiating them from interferences of background structures.
Yang et al. \cite{yang2017deep} proposed a
multi-task deep learning architecture that involves learning of the binary rain streak map, the appearance of rain streaks, and the clean background under one framework. These CNN structures still shown limitations for opaque and heavy rain occlusions. 

\begin{figure*}
	\begin{center}
		\includegraphics[width=0.96\linewidth]{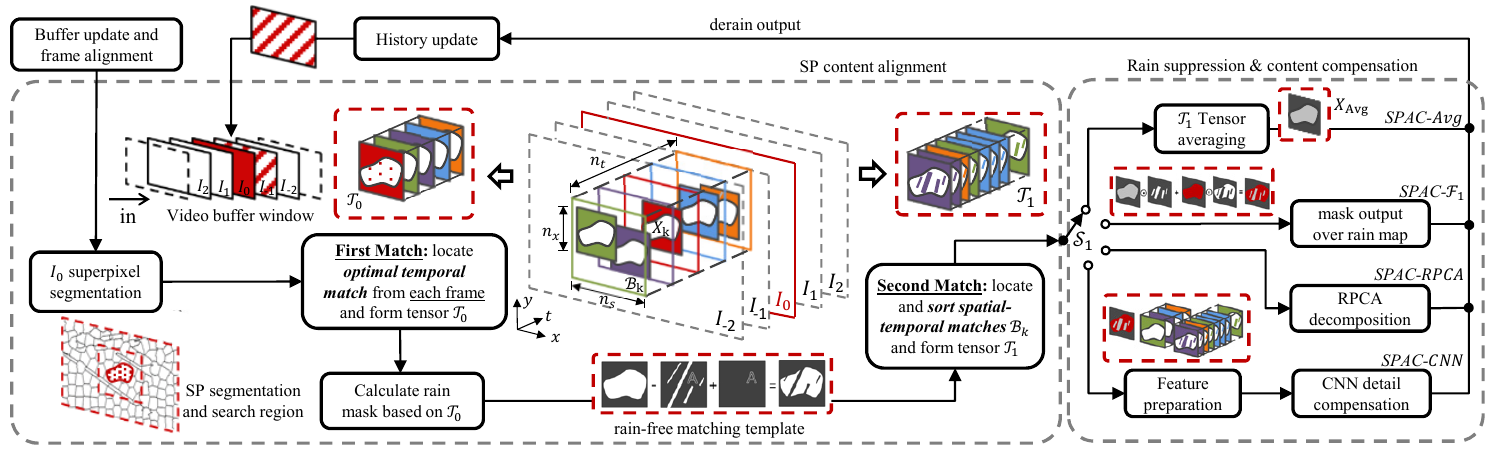}
	\end{center}
	\vspace{-0.5cm}
	\caption{System diagram for the proposed \textit{SPAC} framework. The switch $\mathcal{S}_1$ changes algorithms between \textit{SPAC-Avg}, \textit{SPAC-$\mathcal{F}_1$, \textit{SPAC-RPCA}, and \textit{SPAC-CNN}.}}
	\vspace{-0.4cm}
	\label{fig_system}
\end{figure*}
%%%%%%%%%%%%%%%%%%%%%%%Figure%%%%%%%%%%%%%%%%%%%%%%%

\subsection{Video-based methods}
Most video-based methods rely primarily on temporal analysis of pixel values.
In principle, the intensity of each pixel is analyzed over time; elevated intensity values indicate the presence of rain.
These rain pixels are then restored using pixel values from neighboring frames.
The issue with temporal analysis is that camera motion and dynamic scene motion could also cause intensity spikes, creating false positives.
Thus, methods have been explored to identify these false positives from rain-induced spikes.

For example, Garg and Nayar \cite{Garg2004,Garg2007} proposed to detect candidate rain pixels by thresholding intensity differences between the target frame and its two immediate neighbors. 
True rain pixels are differentiated from false positives by comparing intensity changes to background intensities, and further refined by comparing rain patterns with those of neighboring frames. 
The rain pixel is then restored from the average of the two neighboring frames.
Later works proposed different methods to identify true rain pixels from false positives.
Zhang et al. \cite{zhang2006} checked whether intensity changes of RGB color channels are similar, Bossu et al. \cite{bossu2011rain} applied geometric constraints, such as size and orientation, to candidate rain streaks, and Tripathi and Mukhopadhyay \cite{Tripathi2012} modeled intensity fluctuations in a local spatio-temporal area.
In general, when rain pixels overlap with moving objects, which can generate false positives, this approach does not perform well.
Chen and Chau \cite{chen2014rain} proposed a solution by segmenting motion regions and de-raining them with alternative approaches.
%Chen and Chau \cite{chen2014rain} segment motion regions and apply spatial filtering on them instead of temporal filtering. 

More recently, Jiang et al. \cite{jiang2017novel} decomposed a video into background and rain layers by enforcing a set of priors, such as low-rankedness of background, sparsity of rain, and total variation of rain and background in different directions.
Wei et al. \cite{wei2017should} decomposed a video into background, rain, and foreground (moving objects) layers by modeling the rain layer as patch-based mixture of Gaussians, background layer as a low rank structure, and foreground as a sparse and spatial-temporally continuous structure.
However, the assumption that the background is low-ranked is invalidated when the camera is in motion.
Ren et al. \cite{ren2017video} made a similar matrix decomposition, where background is constrained to be low-ranked, and foreground and rain layers are constrained to be sparse. 
Moving objects are de-rained by temporally aligning them using patch matching, while the moving camera effect is modeled using global frame transform.
Although global frame transform can compensate for camera motion, it is inadequate when camera motion is fast or when the scene contains large depth ranges.
Furthermore, a relatively large number of frames is required for low rank decomposition, which makes their alignment difficult to achieve under fast camera motion.

Kim et al. \cite{kim2015video} estimated a hybrid warped frame from two immediate neighbors using optical flow. This frame, assumed to be rain-free, is compared with the target frame to detect rain candidates, which are then refined using sparse coding. Rain pixels are recovered using block-based matrix completion.
%However, it is challenging to accurately extract this hybrid warped frame. 
There are also works with less emphasis on using temporal data.
Barnum et al. \cite{Barnum2009Frequency} modeled rain streaks in frequency space, Santhaseelan and Asari\cite{santhaseelan2015utilizing} use phase congruency on frame difference images to detect candidate rain streaks, and Kim et al. \cite{kim2015mutli} applied horizontal directional thresholding to detect candidate rain streaks.

In general, video-based methods potentially perform better than image-based methods, since temporal information is available.
However, unless accounted for, camera motion and scene object motion invalidate temporal information and thus affect performance.
Image-based methods, as do some video-based methods, rely on priors or assumptions about properties of rain.
However, as shown in our real world experimental videos, rain could appear in irregular sizes, shapes, brightness, opacity, or distributions; these rain are not easily removed by existing image-based methods.

\section{Proposed Model} \label{sec_proposedModel}

Throughout the paper, scalars are denoted by italic lower-case letters, 2D matrices by upper-case letters, and 3D tensors and functions by script letters.

Fig. \ref{fig_system} shows the system diagram for the proposed rain removal framework based on SuperPixel content Alignment and Compensation (SPAC). The framework can be divided into two parts: first, video content alignment is carried out based on SP units. For each SP, we perform two SP template matching operations, which produce two output tensors: the \textbf{\textit{optimal temporal match tensor}} $\mathcal{T}_0$, and the \textbf{\textit{sorted spatial-temporal match tensor}} $\mathcal{T}_1$. %An intermediate derain output $X_\text{avg}$ is calculated by averaging the \textit{slices}\footnote{A slice is a two-dimensional section of a higher dimensional tensor, defined by fixing all but two indices \cite{kolda2009tensor}.} of the tensor $\mathcal{T}_1$. 
Second, rain removal and scene content compensation are carried out over the slices\footnote{A slice is a two-dimensional section of a higher dimensional tensor, defined by fixing all but two indices \cite{kolda2009tensor}.} of the two tensors $\mathcal{T}_0$ and $\mathcal{T}_1$. Various classical and novel methods such as Robust Principle Component Analysis \cite{Peng2012RASL} (RPCA) and Convolutional Neural Networks \cite{krizhevsky2012imagenet} (CNN),   % , i.e., \textit{SPAC}-$\mathcal{F}_1$, \textit{SPAC-RPCA}, \textit{SPAC-Avg}, and \textit{SPAC-CNN}, 
will be investigated and compared under our proposed \textit{SPAC} framework \cite{chen2018robust} for their respective advantages in the task of scene content restoration.  

\subsection{Robust Content Alignment via Superpixel Spatial-Temporal Matching}

Given a target frame $I_0$ of a rainy video, we consider its immediate past and future neighbor frames to create a sliding buffer window of length $n_t$: 
$\{I_i|i=[-\frac{n_t-1}{2},\frac{n_t-1}{2}]\}$. Here, negative and positive $i$ indicate past and future frames, respectively. We only derain the Y luminance channel. The derain output is used to update the corresponding frame in the buffer to improve rain removal of future target frames (As shown Fig. \ref{fig_system}). This history buffer update mechanism contributes to cleaner derain output especially under heavy rainfall scenarios. 

One of the most important procedures for video-based derain algorithms are the estimation of content correspondence between video frames. With accurate content alignment, rain occlusions could be easily detected and removed with information from the temporal axis. The alignment, however, is no trivial task. Camera motion changes the scene significantly, and the change of perspective causes scene contents at different distance to shift at different parallax. Additionally, scene contents could have their own rigid/non-rigid motion, not to mention the serious interference caused by rain streak occlusions. All these factors make content alignment extremely challenging for rainy videos.

\subsubsection{Content Alignment: Global vs. Superpixel} \label{sec_alignment}

%%%%%%%%%%%%%%%%%%%%%%%Figure%%%%%%%%%%%%%%%%%%%%%%%
\begin{figure}[t]
	\begin{center}
		\includegraphics[width=0.96\linewidth]{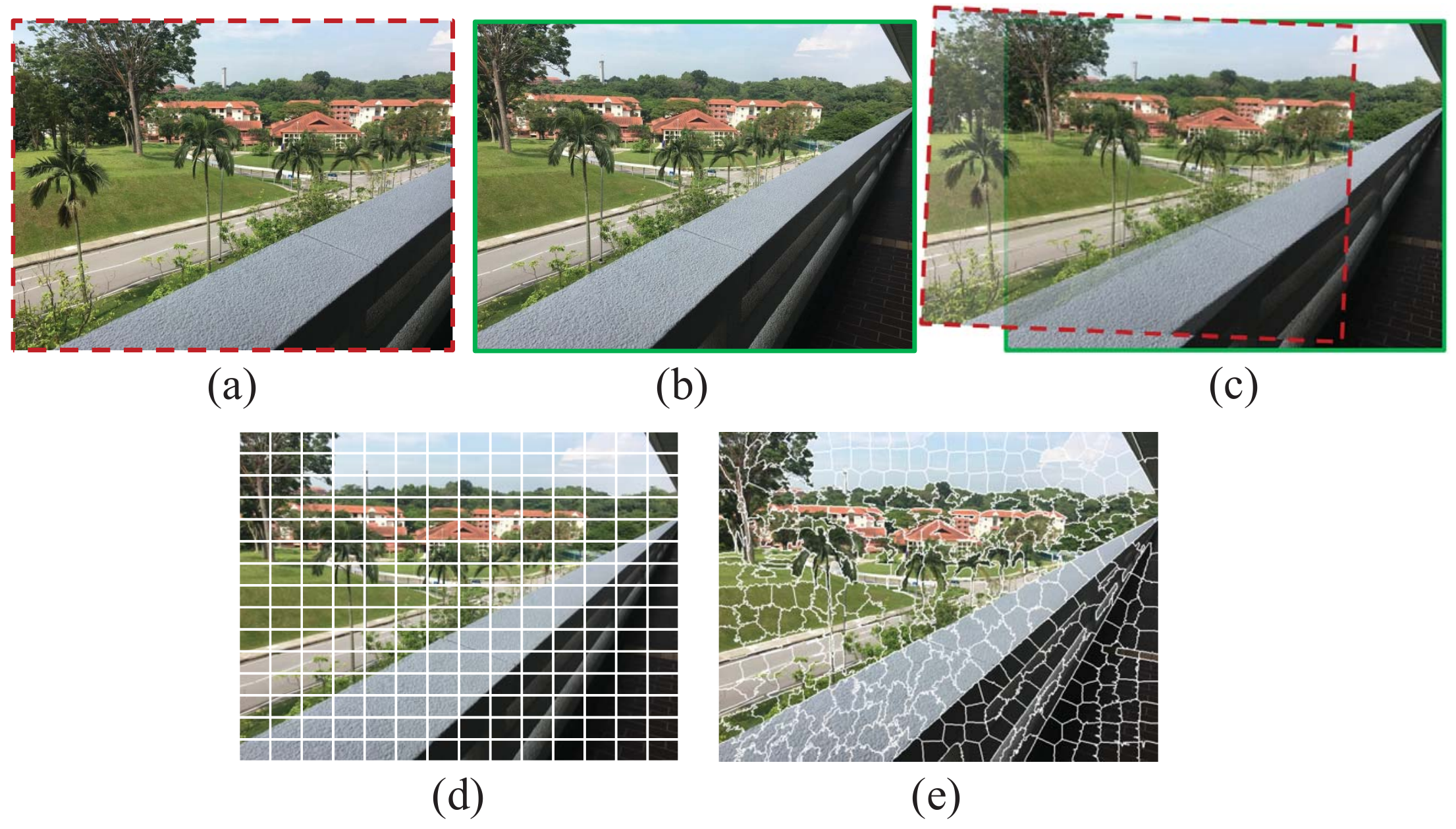}
	\end{center}\vspace{-0.5cm}
	\caption{Motivation of choosing SP as basic processing unit. (c) shows the outcome of geometrical alignment between (a) and (b). (d) and (e) compare segmentation based on rectangular and SP units.}
	\label{fig_warpingOutput}
	\vspace{-0.5cm}
\end{figure}
%%%%%%%%%%%%%%%%%%%%%%%Figure%%%%%%%%%%%%%%%%%%%%%%%

The popular solution to compensate camera motion between two frames is via a homography transform matrix estimated based on global consensus of a group of matched feature points \cite{bay2008speeded, torr2000mlesac}. Due to the reasons analyzed in Sec. \ref{sec_introduction}, perfect content alignment can rarely be achieved for all pixels with a global transform at whole frame level, especially for dynamic scenes with large depth range.
To illustrate, two frames in Fig. \ref{fig_warpingOutput}(a) and (b) are geometrically aligned via global homography transform, and the aligned frames are overlapped and shown in Fig. \ref{fig_warpingOutput}(c). It can be seen that scene contents at far and middle distance ranges are well-aligned; however, for nearer contents (e.g., the banisters), obvious mis-alignment appear. 

The solution naturally turns to pixel-level alignment, which faces no fewer challenges: first, feature points are sparse, and feature-less regions are difficult to align; more importantly, rain streak occlusions will cause serious interferences to feature matching at single pixel level. Contextual information from neighboring area is indispensable to overcome rain interferences. This leads us to our solution: to decompose images into smaller depth consistent units.

The concept of SuperPixel (SP) is to group pixels into perceptually meaningful atomic regions \cite{achanta2012slic,bergh2012seeds,li2015superpixel}. Boundaries of SP usually coincide with those of the scene contents. Comparing Fig. \ref{fig_warpingOutput}(d) and (e), the SPs are very adaptive in shape, and are more likely to segment uniform depth regions compared with rectangular units.
We adopt SP as the basic unit for content alignment.

% \textcolor{Green}{In our implementation, we use the SLIC superpixel segmentation method \cite{achanta2012slic}, an iterative regional pixel clustering algorithm that clusters each pixel to an initiated center grid according to the their respective normalized spatial and color distances. The number of SPs created is a parameter that can be adjusted depending on the target application and the scene captured by the input video. Other SP segmentation algorithms, such as \cite{bergh2012seeds} and \cite{li2015superpixel}, can also be used.}

\subsubsection{Optimal Temporal Matching for Rain Detection}\label{sec_T0}

Let $\mathcal{P}_k$ denote the set of pixels that belong to the $k$-th SP on $I_0$. Let $X_k\in\mathbb{R}^{n_x\times n_x}$ be the bounding box that covers all pixels in $\mathcal{P}_k$ ($\mathcal{P}_k\subset X_k$). Let $\mathcal{B}_{k}\in\mathbb{R}^{n_\text{s}\times n_\text{s}\times n_t}$ denote a spatial-temporal buffer centered on $\mathcal{P}_k$. As illustrated in Fig. \ref{fig_system}, the temporal range of $\mathcal{B}_k$ spans the entire sliding buffer window, and its spatial range $n_s\times n_s$ is set to cover the possible motion range of $\mathcal{P}_k$ in its neighboring frames.

Pixels within the same SP are very likely to belong to the same object and show identical motion between adjacent frames.
Therefore, we can approximate the SP's appearance in other frames based on its appearance in the current frame via linear translations. 

\begin{figure}
	\centering
	\includegraphics[width=0.76\linewidth]{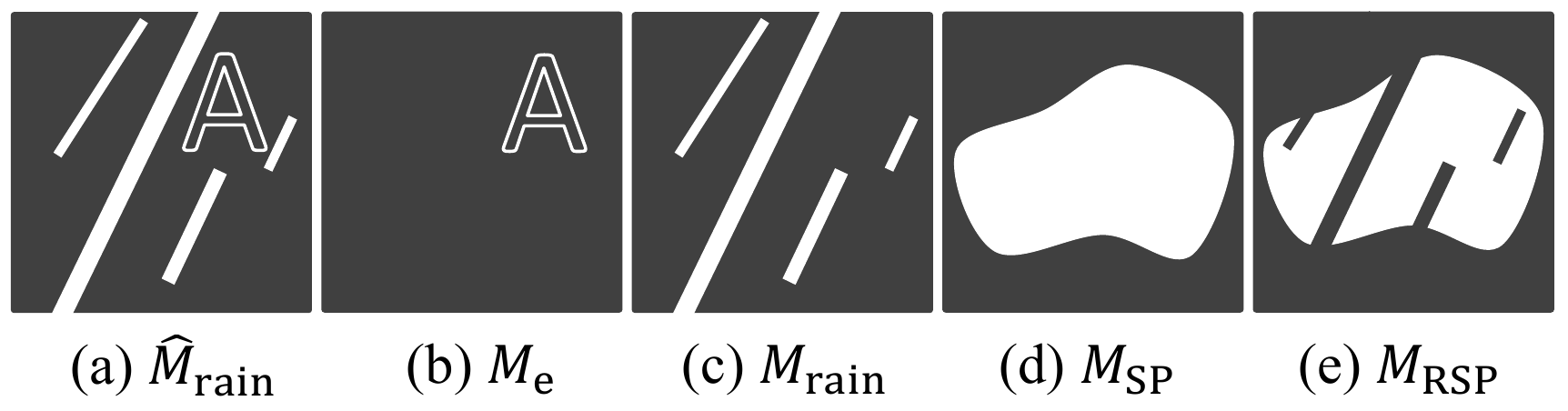}
	\caption{Illustration of various masks and matching templates used in the proposed algorithm.}
	\label{fig_maskMatch}
\end{figure}

Searching for the reference SP is done by template matching of the target SP at all candidate locations in $\mathcal{B}_k$. A match location is found at frame $I_{t'}$ according to: 
\begin{align}\label{eqn_temporalSearch}
(\hat{u},\hat{v})=\argmin_{u,v}\sum_{(x,y)\in X_k}&|\mathcal{B}_k(x+u,y+v,t')\\\notag 
&-X_k(x,y)|^2\odot M_\text{SP}(x,y).
\end{align}
As shown in Fig.~\ref{fig_maskMatch}(d), $M_\text{SP}$ indicates SP pixels $\mathcal{P}_k$ in the bounding box $X_k$. $\odot$ denotes element-wise multiplication. Each match at a different frame becomes a \textit{temporal slice} for the \textbf{\textit{optimal temporal match tensor}} $\mathcal{T}_0\in \mathbb{R}^{n_x\times n_x\times n_t}$: \vspace{-0.2cm}
\begin{equation}\label{eqn_T0}
\mathcal{T}_0(\cdot,\cdot,t')= \mathcal{B}_k(x+\hat{u},y+\hat{v},t'),~(x,y)\in X_k.
\end{equation}
% here $\mathcal{T}_0(\cdot,\cdot,t')$ denotes the $t'$th slice of the tensor $\mathcal{T}_0$.

%%%%%%%%%%%%%%%%%%%%%%%Figure%%%%%%%%%%%%%%%%%%%%%%%
%%%%%%%%%%%%%%%%%%%%%%%Figure%%%%%%%%%%%%%%%%%%%%%%%

Based on the temporal clues provided by $\mathcal{T}_0$, a rain mask can be estimated. Since rain increases the intensity of its covered pixels \cite{Garg2004}, rain pixels in $X_k$ are expected to have higher intensity than their collocated temporal neighbors in $\mathcal{T}_0$. We first compute a binary tensor $\mathcal{M}_{\text{0}}\in\mathbb{R}^{n_x\times n_x\times n_t}$ to detect positive temporal fluctuations: \vspace{-0.2cm} 
\begin{align}
\mathcal{M}_{0}&=\begin{cases}
1& \mathcal{R}(X_k,n_t)- \mathcal{T}_0 \ge \epsilon_{\text{rain}}\\
0& \text{otherwise}
\end{cases},
\end{align}
where operator $\mathcal{R}(\Phi,\psi)$ is defined as replicating the 2D slices $\Phi\in\mathbb{R}^{n_1\times n_2}$ $\psi$ times and stacking along the thrid dimension into a tensor of $\mathbb{R}^{n_1\times n_2\times \psi}$. To robustly handle re-occurring rain streaks, we classify pixels as rain when at least 3 positive fluctuations are detected in $\mathcal{M}_0$. An initial rain mask $\hat{M}_{\text{rain}}\in\mathbb{R}^{n_x\times n_x}$ can be calculated as: \vspace{-0.2cm}
\begin{equation}
\hat{M}_{\text{rain}}(x,y)=[\sum_t \mathcal{M}_{\text{0}}(x,y,t)] \ge 3.
\end{equation}\vspace{-0.3cm}

Due to possible content mis-alignment, edges of background could be misclassified as rain. Since rain steaks don't affect values in the chroma channels (Cb and Cr), a rain-free edge map $M_{\text{e}}$ could be calculated by thresholding the sum of gradients of these two channels with $\epsilon_\text{e}$. The final rain mask $M_{\text{rain}}\in\mathbb{R}^{n_x\times n_x}$ is calculated as:\vspace{-0.2cm}
\begin{equation}
M_{\text{rain}} = \hat{M}_{\text{rain}}\odot (1-M_e).
\end{equation}
A conceptual demonstration of $\hat{M}_{\text{rain}}$, $M_{\text{e}}$, and $M_{\text{rain}}$ is shown in Fig. \ref{fig_maskMatch}(a), (b), and (c), respectively. In our implementation, $\epsilon_\text{rain}$ is set to $0.012$ while $\epsilon_\text{edge}$ is set to $0.2$.

\subsubsection{Sorted Spatial-Temporal Template Matching for Rain Occlusion Suppression} \label{sec_T1}

\begin{figure}
	\centering
	\includegraphics[width=1\linewidth]{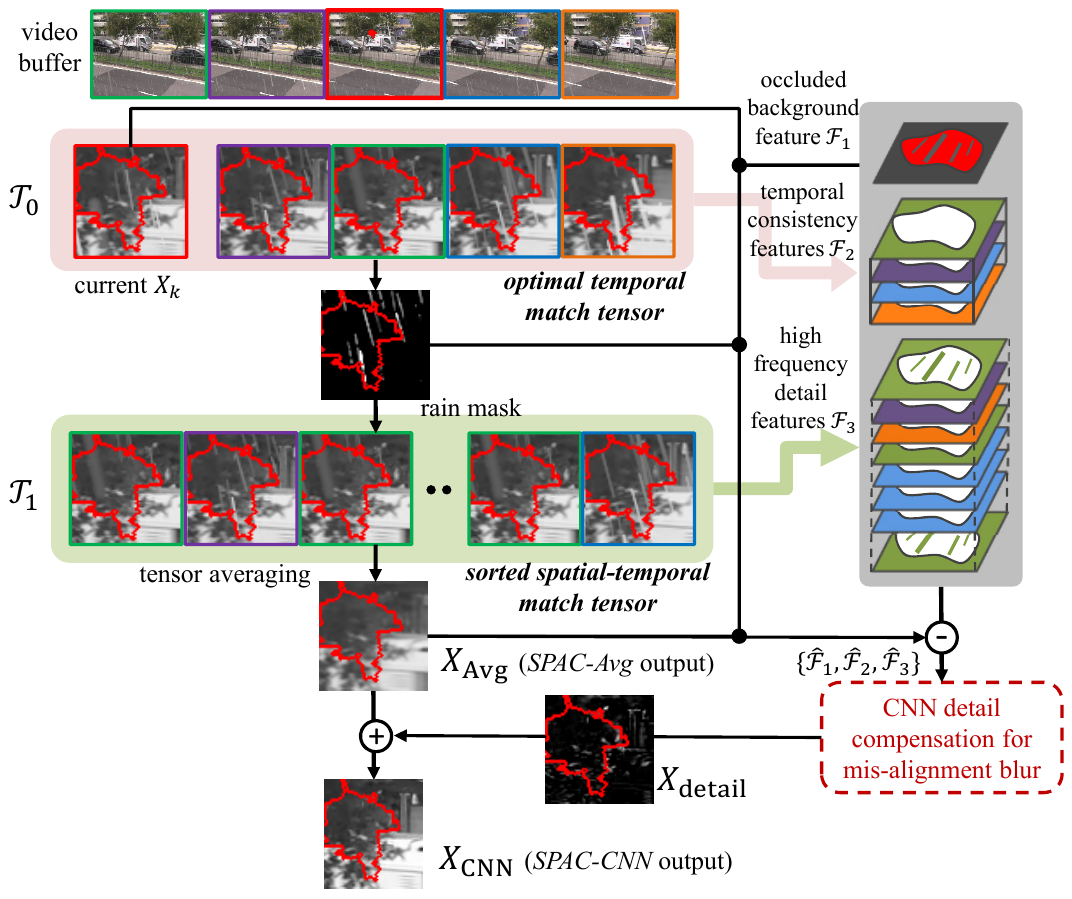}
	\caption{Illustration of calculation flow for \textit{SPAC-Avg} and feature preparation for \textit{SPAC-CNN}. }
	\label{fig_T0T1forming}
\end{figure}

A second round of template matching will be carried out based on the following cost function: \vspace{-0.2cm}
\begin{align}\label{eqn_T1cost}
E(u,v,t)= \sum_{(x,y)\in X_k}&|\mathcal{B}_k(x+u,y+v,t)\\ \notag  
- &X_k(x,y)|^2\odot M_{\text{RSP}}(x,y).
\end{align}
The \textbf{rain-free matching template} $M_
\text{RSP}$ is calculated as: \vspace{-0.2cm} 
\begin{equation}
M_{\text{RSP}}= M_{\text{SP}} \odot (1-M_{\text{rain}}).
\end{equation}
As shown in Fig. \ref{fig_maskMatch}(e), only the \textbf{\textit{rain-free background SP pixels}} will be used for matching. 
Candidate locations in $\mathcal{B}_k$ (except current frame $\mathcal{B}_k(\cdot,\cdot,0)$) are sorted in ascending order based on their cost $E$ defined in Eq. (\ref{eqn_T1cost}). The top $n_{st}$ candidates with smallest $E$ will be stacked as slices to form the \textbf{\textit{sorted spatial-temporal match tensor}} $\mathcal{T}_1\in\mathbb{R}^{n_x\times n_x\times n_{st}}$.
With masked template matching eliminating the interferences from rain, the slices of $\mathcal{T}_1(\cdot,\cdot,t)$ are expected to be well-aligned to the current target SP.
Since rain pixels are temporally randomly and sparsely distributed within $\mathcal{T}_1$, when $n_{st}$ is sufficiently large, we can get a good estimation of the rain free image through \textit{tensor slice averaging}, which functions to suppress rain induced intensity fluctuations, and recover the occluded background pixels: \vspace{-0.2cm}
% When $n_{st}$ is sufficiently large, along the temporal dimension of $\mathcal{T}_1$, the rain occurring frequency will be statistically stable for all pixels. Therefore, if we average the temporal slices of $\mathcal{T}_1$, the outcome is supposed to washout rain streak occlusions, and stably bring out the occluded pixels, which provides a good estimation of rain free image: \vspace{-0.2cm}
\begin{equation}\label{eqn_Xavg}
X_\text{Avg}=\frac{\sum_{t}\mathcal{T}_1(\cdot,\cdot,t)}{n_{st}}.
\end{equation}
If we only extract one tensor slice from each frame as done for $\mathcal{T}_0$, there will be too few samples for suppression of heavy rain.
For $\mathcal{T}_1$, we set $n_{st} > n_t$; thus, $\mathcal{T}_1$ contains corresponding scene content from neighboring frames along with their slightly shifted copies.
%Since there are only four neighboring frames, which is too few for suppression of heavy rain, we set $n_{st} > n_t$.
%$\mathcal{T}_1$ thus contains corresponding scene content from neighboring frames along with their slightly shifted copies.
Hence, $\mathcal{T}_1$ tensor averaging is similar to weighted spatio-temporal averaging of aligned scene content, where the binary weights are dependent on similarity to the target patch.
We denote this method as \textit{SPAC-Avg}, and use it as the baseline method in our \textit{SPAC} framework.
Fig.\ref{fig_T0T1forming} illustrates the steps involved in computing \textit{SPAC-Avg}.

%\textcolor{Gray}{
%As will be visually validated in Sec. \ref{sec_exp_veiling&Accu}, Tensor averaging in Eq. (\ref{eqn_Xavg}) proves to be efficient in suppressing not only occluding rain streaks, but also veiling rain that exhibits semi-transparent intensity fluctuations and distortions in a large area (COMMENT:the section does not show any visual results for spac-avg). $X_\text{Avg}$ already provides a reasonable derain output, which we denote as \textit{SPAC-Avg}. 
%Fig. \ref{fig_T0T1forming} gives a visual illustration of the calculation procedures of \textit{SPAC-Avg}.}

%%%%%%%%%%%%%%%%%%%%%%%Figure%%%%%%%%%%%%%%%%%%%%%%%
%%%%%%%%%%%%%%%%%%%%%%%Figure%%%%%%%%%%%%%%%%%%%%%%%

\subsection{Rain Removal and Low Rank Content Modeling via Robust Principle Component Analysis}
%\subsubsection{Modeling of Tensor $\mathcal{T}_1$ as \textit{low-rank} matrix}
%\textcolor{Gray}{
%The \textit{sorted spatial-temporal match tensor} $\mathcal{T}_1\in\mathbb{R}^{n_x\times n_x\times n_{st}}$ consists of $n_{st}$ slices of well-aligned image patches. However, since small mis-alignments between the slices are unavoidable (especially when camera motion is large and scene content is dynamic), direct averaging of $\mathcal{T}_1$ slices in Eq. \ref{eqn_Xavg} creates noticeable blur. 
%(COMMENT: the mis-alignment and the resulting blur is not 'unavoidable'. It is intended so that rain can be cleanly removed. Furthermore, "especially when camera motion is large and scene content is dynamic" (above sentence) implies that we are going to propose a new method- RPCA - that works better when camera motion is large.
%}

As shown in experimental results, although \textit{SPAC-Avg} proves to be very effective in suppressing rain, it also blurs background details. 
Hence, in this section, we attempt to use a more advanced mathematical model -- Robust Principle Component Analysis (RPCA) \cite{Candes2011robust}.

%\textcolor{Gray}{
%In this section, we attempt to use a more advanced mathematical model -- Robust Principle Component Analysis (RPCA) \cite{Candes2011robust} -- to model the background, which has been proven to be more robust against corrupted outliers. (COMMENT: RPCA is not a background model, but a model with background+sparse outliers)
%}

%RPCA has been used in many applications to recover corrupted entries from data \cite{yong2017robust}, where the collection of data is low ranked and the corrupted entries are sparse. Its main advantage over averaging-based methods is that it is robust to corruptions of large magnitudes.

RPCA was originally proposed to recover corrupted entries from data, where the collection of data is low ranked and the corrupted entries are sparse. 
However, it has been successfully applied to many other applications, such as \cite{yong2017robust, gao2014block, dai2014simple}. 
Its main advantage over averaging-based methods is that it is robust to corruptions of large magnitudes.

%The \textit{sorted spatial-temporal match tensor} $\mathcal{T}_1\in\mathbb{R}^{n_s\times n_s\times n_{st}}$ consists of $n_{st}$ slices of well-aligned image patches.
Let $\mathcal{V}(\cdot):\mathbb{R}^{n_x\times n_x}\rightarrow \mathbb{R}^{n_p\times 1}$ denote the operator that extracts the pixels of a target SP from a $n_x\times n_x$ 2D rectangular patch, and then stacks them as a vector $n_p\times 1$. Here, $n_p~(n_p\leq n_x\times n_x)$ denotes the number of pixels in the target SP. A 2D matrix $\Psi\in\mathbb{R}^{n_p\times n_{st}}$ can be formed from the target SP and slices in $\mathcal{T}_1$:
\begin{equation}
%\Psi\dot{=}[\mathcal{V}(\mathcal{T}_1(\cdot,\cdot,1))|\mathcal{V}(\mathcal{T}_1(\cdot,\cdot,2))|...|\mathcal{V}(\mathcal{T}_1(\cdot,\cdot,n_{st-1}))].
\Psi\dot{=}[X_k \;|\; \mathcal{V}(\mathcal{T}_1(\cdot,\cdot,1)) \;|\; ... \;|\; \mathcal{V}(\mathcal{T}_1(\cdot,\cdot,n_{st-1}))].
\end{equation}
Should the SP contents in $\mathcal{T}_1$ be aligned, columns of $\Psi$ will be similar with each other, and $\Psi$ should be low-rank. 
%\subsubsection{Modeling of Rain Occlusion as Sparse Errors}
%\textcolor{Gray}{(COMMENT: better to discuss this in results section?)
%However, the low-rank structure of $\Psi$ can be easily violated due to the presence of rain streak occlusion. Additionally, misalignment caused by camera/object motion, and changing imaging conditions etc., also introduces errors. }
We model the rain-free content of $\Psi$ as a \textit{low-rank} matrix $L$; the rain occlusions and small shifts typically affect only a small fraction of pixels in an image, therefore we model them as a sparse matrix $R$.

Finally, the rain removal problem can now be expressed as:
\begin{equation}\label{eqn_nonconvex}
\min_{L,R}{\text{rank}(L)+\gamma||R||_0},~s.t.~\Psi=L+R. 
\end{equation}
%where $X_{rain}= [\text{vec}(e_i)|...|\text{vec}(e_{n_{st}})]$. 
Here, the $\ell_0$-norm $||\cdot||_0$ counts the number of nonzero entries in the matrix $R$. $\gamma$ is a parameter that trades off the rank of $L$ versus the sparsity of $R$. 

The problem in Eq. (\ref{eqn_nonconvex}) is highly nonconvex due to the calculation of matrix rank and $\ell^0$-norm. As suggested in \cite{Candes2011robust}, we relax it to its convex surrogate as:
\begin{equation}\label{eqn_convex}
\min_{L,R}{||L||_*+\gamma||R||_1},~s.t.~\Psi=L+R,
\end{equation}
which replaces  $\text{rank}$ with the \textit{nuclear norm} of $L$ (which equals the sum of its singular values), and replaces the $\ell^0$-norm $||R||_0$ with the $\ell^1$-norm: $\sum_{i,j}|\Psi_{i,j}|$. It has been shown in \cite{Candes2011robust} that Eq. (\ref{eqn_convex}) can exactly recover $L$ in Eq. (\ref{eqn_nonconvex}) as long as long as the rank of the matrix $\Psi$ is not too high and the number of nonzero entries in $R$ is not too large. This condition is generally met when the alignment quality of $\mathcal{T}_1$ is high, and when the rain occlusions are not too dense.
%(\textcolor{Gray}{yet as mentioned earlier, "misalignments are unavoidable". It is not clear that with unavoidable misalignments, alignment quality can still be high})

The convex problem in Eq. (\ref{eqn_convex}) can be efficiently solved via modern first-order optimization techniques such as the Augmented Lagrange Multiplier (ALM) algorithm \cite{bertsekas1999nonlinear}. 
The derained output SP $X_\text{LR}$ is extracted from the first column of \text{low-rank} matrix $L$ using the inverse of $\mathcal{V}(\cdot)$.
We denote this method as \textit{SPAC-RPCA}.
%\textcolor{Gray}{
%The column in the \text{low-rank} matrix $L$ that corresponds to the current frame will be put back to its original locations in the target SP as the derain output $X_\text{LR}$ could be better written, not sure how yet, which we denote as the output of the method \textit{SPAC-RPCA}. }

%\textcolor{Gray}{(COMMENT: averaging (SPAC-Avg) is equivalent to standard PCA, not RPCA) It is worth noting that \textit{SPAC-Avg} is actually one solution of Eq. (\ref{eqn_nonconvex}), however with a harder constraint that $\text{rank}(L)=1$.} 
%\textcolor{Gray}{\textit{SPAC-RPCA} can model the sparse errors more robustly, which gives tolerance to small content mis-alignments. Therefore, \textit{SPAC-RPCA} is supposed to preserve the background details better as compared with \textit{SPAC-Avg}.}

%\textcolor{blue}{
%To properly model a low ranked background, a large number of data samples is required. 
%VST models each frame of a video sequence as a data sample and uses all frames as its background.
%Due to lack of neighbor frames. Our method uses $n_st$ instead of $n_t$.
%Although the samples in $n_st$ contains shifted copies, we show in the experimental results that SPAC-RPCA is tolerant to such shifts. 
%}

% In our implementation, $\lambda$ is set to $1/\max(\sqrt{n_p},\sqrt{n_{st}})$, which we found to be able to achieve a good balance between rain removal and background detail preservation.

\subsection{Rain Removal and Content Compensation in a CNN framework}

%In Eq. (\ref{eqn_Xavg}), the averaging of $\mathcal{T}_1$ slices (\textit{SPAC-Avg}) provides a good estimation of rain free image; however, it creates noticeable blur due to un-avoidable mis-alignment, especially when the camera motion is fast. 
As to be shown in Sec. \ref{sec_experiments}, when camera motion gets faster (which leads to larger mis-alignment errors), and when rain occlusions get dense, \textit{SPAC-RPCA} shows limitations. Under both circumstances, the low-rank assumption of $L$, and the sparseness assumption of $R$ are violated.

We further investigate another solution for this challenging scenario with a CNN model, which is expected to have a better capacity to robustly model the rain-free background contents with its deep and highly non-linear layers. 
In this section, we aim at designing a CNN structure (denoted as \textit{SPAC-CNN}) for restoring the lost high frequency scene content details back to $X_\text{Avg}$, however without re-introducing the rain streaks (rain streaks are also high frequency signals).
We will first introduce some informative features for this task, then we proceed to the network structure design details.
% The input features as well as the CNN architecture details will be explained in this section.

\subsubsection{Occluded Background Feature} 

% As has been analyzed in Sec. \ref{sec_T1}, the averaging of $\mathcal{T}_1$ slices suppresses the rain intensity fluctuations and brings out occluded background pixels, therefore 
$X_\text{Avg}$ from Eq. (\ref{eqn_Xavg}) can be used as one important clue to recover rain occluded pixels. Rain streak pixels indicated by the rain mask $M_\text{rain}$ are replaced with corresponding pixels from $X_\text{Avg}$ to form the first feature $\mathcal{F}_1\in\mathbb{R}^{n_x\times n_x\times 1}$:\vspace{-0.2cm}
\begin{equation}\label{eqn_feature1}
\mathcal{F}_1= X_k\odot (1-M_\text{rain})+ X_\text{Avg}\odot M_\text{rain}.
\end{equation}

Feature $\mathcal{F}_1$ itself is already a good derain output, which we denote as \textit{SPAC-$\mathcal{F}_1$}. However its quality is greatly limited by the correctness of the rain mask $M_\text{rain}$. For \textbf{false positive}\footnote{False positive rain pixels refer to background pixels falsely classified as rain; false negative rain pixels refer to rain pixels falsely classified as background.} rain pixels, $X_\text{Avg}$ will introduce content detail loss; for \textbf{false negative} pixels, rain streaks will be added back from $X_k$. This calls for more informative features for the CNN.

\subsubsection{Temporal Consistency Feature}

The temporal consistency feature is designed to handle \textbf{false negative} rain pixels in $M_\text{rain}$, which falsely add rain streaks back to $\mathcal{F}_1$. For a correctly classified and recovered pixel (a.k.a. \textbf{true positive}) in Eq. (\ref{eqn_feature1}), intensity consistency should hold such that for the collocated pixels in the neighboring frames, there are only positive intensity fluctuations caused by rain in \textit{those} frames. Any obvious negative intensity drop along the temporal axis is a strong indication that such pixel is a \textbf{false negative} pixel.

The temporal slices in $\mathcal{T}_0$ establishes optimal temporal correspondence at each frame, which embeds enough information for the CNN to deduce the above analyzed false negative logic, therefore they shall serve as the second feature $\mathcal{F}_2\in\mathbb{R}^{n_x\times n_x\times (n_t-1)}$:
\begin{equation}
\mathcal{F}_2= \{\mathcal{T}_0(\cdot,\cdot,t)|~t=[-\frac{n_t-1}{2},\frac{n_t-1}{2}],~t\neq 0\}.
\end{equation}\vspace{-0.5cm}

\subsubsection{High Frequency Detail Feature} 

The matched slices in $\mathcal{T}_1$ are sorted according to their rain-free resemblance to $X_k$, which provide good reference to the content details with supposedly small mis-alignment. We directly use the tensor $\mathcal{T}_1$ as the last group of features $\mathcal{F}_3= \mathcal{T}_1\in\mathbb{R}^{n_x\times n_x\times n_{st}}$. This feature will compensate the detail loss introduced by the operations in Eq. (\ref{eqn_feature1}) for \textbf{false positive} rain pixels.

In order to facilitate the network training, we limit the mapping range between the input features and regression output by removing the low frequency component ($X_\text{avg}$) from these input features. Pixels in $X_k$ but outside of the SP $\mathcal{P}_k$ is masked out with $M_\text{SP}$:\vspace{-0.1cm}
\begin{align}
\hat{\mathcal{F}}_1&= (\mathcal{F}_1- X_\text{Avg})\odot M_\text{SP},\\ \notag
\hat{\mathcal{F}}_2&= (\mathcal{F}_2- \mathcal{R}(X_\text{Avg},n_t-1))\odot\mathcal{R}(M_\text{SP},n_t-1),\\ \notag
\hat{\mathcal{F}}_3&= (\mathcal{F}_3- \mathcal{R}(X_\text{Avg},n_{st}))\odot\mathcal{R}(M_\text{SP},n_{st}).
\end{align}
The final input feature set is $\{\hat{\mathcal{F}}_1,~\hat{\mathcal{F}}_2,~\hat{\mathcal{F}}_3\}$. The feature preparation process for \textit{SPAC-CNN} is summarized in Fig. \ref{fig_T0T1forming}.

\subsubsection{CNN Structure and Training Details}\label{sec_cnn}

%%%%%%%%%%%%%%%%%%%%%%%Figure%%%%%%%%%%%%%%%%%%%%%%%
\begin{figure}[t]
	\begin{center}
		\includegraphics[width=0.96\linewidth]{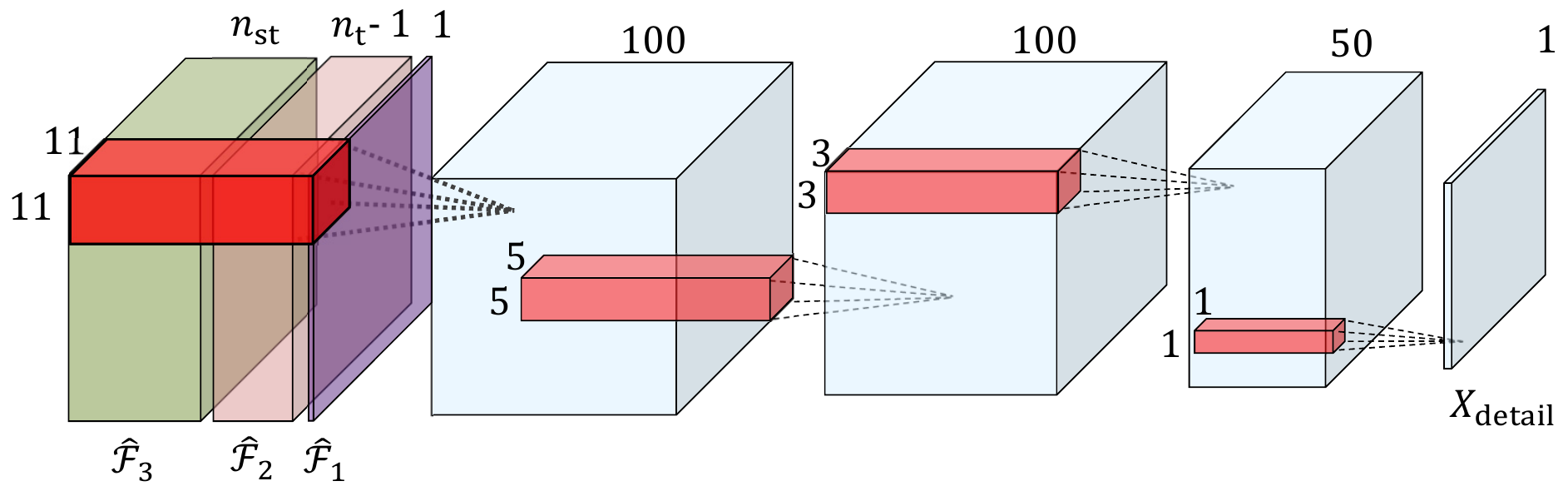}
	\end{center}
	\vspace{-0.5cm}
	\caption{CNN architecture for compensation of mis-alignment blur. Each convolutional layer is followed by a rectified linear unit (ReLU).}
	\vspace{-0.5cm}
	\label{fig_cnnArchitecture}
\end{figure}
%%%%%%%%%%%%%%%%%%%%%%%Figure%%%%%%%%%%%%%%%%%%%%%%%

%%%%%%%%%%%%%%%%%%%%%%%Figure%%%%%%%%%%%%%%%%%%%%%%%
\begin{figure*}[!t]
	\centerline{\subfloat{\includegraphics[width=0.96\linewidth]{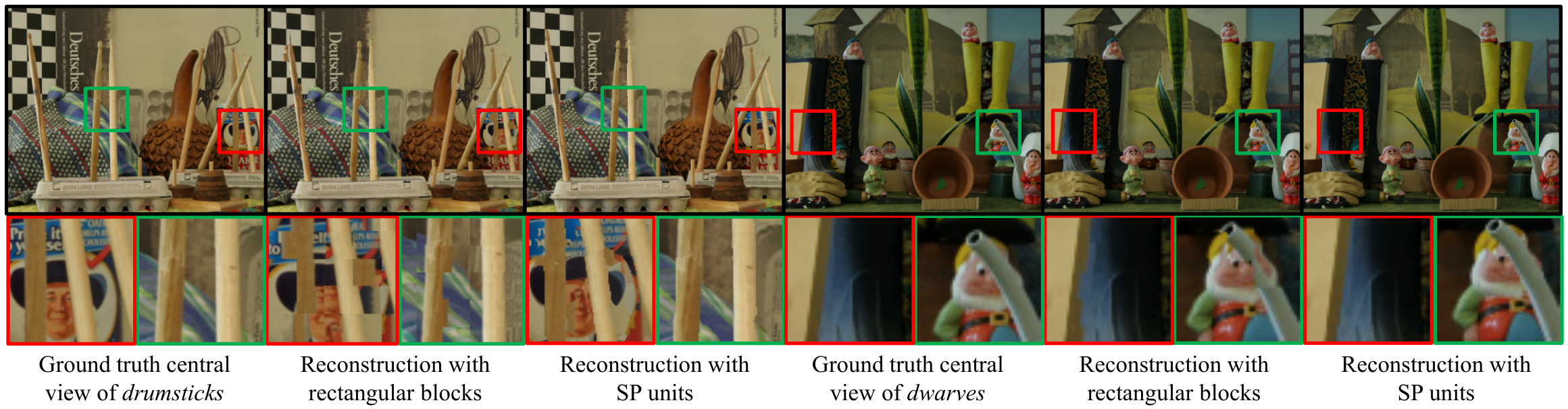}}}
	\caption{Visual comparison of reconstructed central views for the stereo data \textit{drumsticks} and \textit{dwarves} from the Middlebury Stereo Dataset \cite{scharstein2003high}, based on matched rectangular patches, and SP units. Second row show zoom-in details of highlighted regions of the respective first row images.}
	\label{fig_spReconMiddlebury}
\end{figure*}
%%%%%%%%%%%%%%%%%%%%%%%Figure%%%%%%%%%%%%%%%%%%%%%%%

The CNN architecture is designed as shown in Fig. \ref{fig_cnnArchitecture}. The network consists of four convolutional layers with decreasing kernel sizes of 11, 5, 3, and 1. All layers are followed by a rectified linear unit (ReLU). Our experiments show this fully convolutional network is capable of extracting useful information from the input features and efficiently providing reliable predictions of the content detail $X_\text{detail}\in\mathbb{R}^{n_x\times n_x\times 1}$. The final rain removal output will be:\vspace{-0.2cm}
\begin{equation}
X_\text{derain}= X_\text{Avg}+ X_\text{detail}.
\end{equation}

For the CNN training, we minimize the $\mathcal{L}_2$ distance between the derain output and the ground truth scene:
\begin{equation}\vspace{-0.2cm}
E=[\hat{X}- X_\text{Avg}- X_\text{detail}]^2,
\end{equation}
here $\hat{X}$ denotes the ground truth clean image. We use stochastic gradient descent (SGD) to minimize the objective function. Mini-batch size is set as 50 for better trade-off between speed and convergence. The Xavier approach \cite{glorot2010understanding} is used for network initialization, and the ADAM solver \cite{kingma2014adam} is adpatoed for system training, with parameter settings $\beta_1=$ 0.9, $\beta_2=$ 0.999, and learning rate $\alpha=$ 0.0001.

To create the training rain dataset, we first took a set of 8 rain-free VGA resolution video clips of various city and natural scenes. The camera motions are diverse for the dataset, e.g.,  panning slowly with unstable movements, or mounted on a fast moving vehicle with speed up to 30 \textit{km/h}.
Next, rain was synthesized over these video clips with the commercial editing software \textit{Adobe After Effects} \cite{adobeAE}, which can create realistic synthetic rain effect for videos with adjustable parameters such as raindrop size, opacity, scene depth, wind direction, and camera shutter speed. This provides us a diverse rain visual appearances for the network training.

We synthesized 3 to 4 different rain appearances with different synthetic parameters over each video clip, which provides us 25 rainy scenes. For each scene, 21 frames were randomly extracted (together with their immediate buffer window for calculating features). Each scene was segmented into approximately 300 SPs, therefore finally we have around 157,500 patches in the training dataset. 

\section{Performance Evaluation}\label{sec_experiments}

In this section, we evaluate the functional components as well as overall derain performance of the methods under the proposed \textit{SPAC} framework, and compare them with recent state-of-the-art rain removal methods. 

Each frame was segmented into around 300 SPs using the SLIC method \cite{achanta2012slic}. For a VGA resolution frame, this results in average bounding box sizes of $n_x=32$.
By considering two previous and two future neighbor frames, the spatial-temporal buffer $\mathcal{B}_k$ dimension was $n_\text{s}\times n_\text{s}\times n_\text{t}=$ 30$\times$30$\times$5. 
The length of tensor $\mathcal{T}_1$, $n_{st}$, was set to 10. 

For \textit{SPAC-RPCA}, $\lambda$ is set to $1/\max(\sqrt{n_p},\sqrt{n_{st}})$, which we found to be able to achieve a good balance between rain removal and background detail preservation. 
For \textit{SPAC-CNN}, MatConvNet \cite{Vedaldi2015mat} was adopted for model training, which took approximately 54 hours to converge over the training dataset introduced in Sec. \ref{sec_cnn}. The training and all subsequent experiments were carried out on a desktop with Intel E5-2650 CPU, 56GB RAM, and NVIDIA GeForce GTX 1070 GPU.

All the image and video results reported in this section can be accessed online\footnote{Testing dataset and results available at: \url{https://github.com/hotndy/SPAC-SupplementaryMaterials}}.

\subsection{SP Matching Efficiency Evaluation}

%%%%%%%%%%%%%%%TABLE%%%%%%%%%%%%%%%%
{\renewcommand{\arraystretch}{1.2}
	\begin{table}[t]
		\begin{center}
			\small\centering\setlength\tabcolsep{1pt}
			\caption{Central view reconstruction PSNR (\textit{dB}) based on six side views for the Middlebury Stereo Dataset \cite{scharstein2003high}.}
			\label{tbl_matchingPSNR}
			\begin{tabular}{|>{\centering\arraybackslash}m{2cm}|>{\centering\arraybackslash}m{2.4cm}|>{\centering\arraybackslash}m{2.2cm}|>{\centering\arraybackslash}m{1.8cm}|}
				\hline 		
				Data   		&block Matching PSNR 	& SP Matching PSNR	& SP Matching Advantage\\ \hline\hline
				Art 		&30.68			&32.18		&+1.51 \\ \hline
				Books		&34.74			&35.83		&+1.09 \\ \hline
				Computer 	&34.69			&36.40		&+1.71 \\ \hline
				Dolls 		&33.02			&34.50		&+1.48 \\ \hline
				Drumsticks	&29.09			&30.96		&+1.87 \\ \hline
				Dwarves		&35.55			&36.80		&+1.25 \\ \hline
				Laundry		&33.03			&33.75		&+0.72 \\ \hline
				Moebius		&34.05			&35.35		&+1.30 \\ \hline
				Reindeer	&35.15			&35.42		&+0.27 \\ \hline\hline
				\textbf{Average}		&\textbf{33.33}			&\textbf{34.58}		&\textbf{+1.24} \\ \hline
			\end{tabular}\label{tbl_spReconstruct}
		\end{center}
	\end{table}
}
%%%%%%%%%%%%%%%TABLE%%%%%%%%%%%%%%%%

For all methods under the proposed \textit{SPAC} framework, the choice of SP as the basic operation unit is key to their performances. When other decomposition units are used instead (e.g., rectangular patches), matching accuracy deteriorates, and very obvious averaging blur will be introduced especially at object boundaries.

We quantitatively evaluate the advantage of content alignment based on SP units compared with rectangular units over the Middlebury Stereo Dataset \cite{scharstein2003high}. The dataset contains 9 image sequences, with each looking at a target scene from 7 different viewing angles. The baselines between the angles are large enough to simulate fast camera motion. 

A straightforward reconstruction method is used for this evaluation. The 7 views are treated as 7 consecutive video frames and forms a 7-frame buffer. For a certain target matching unit (either SP or rectangular block) on the central view (i.e., the 4th frame), we find a best match in the spatial-temporal buffer with the smallest matching error defined in Eq. (\ref{eqn_T1cost}). Note that in this experiment, $M_\text{RSP}$ is set equal to the SP mask $M_\text{SP}$, since there is no rain in this dataset; and for rectangular patches, the mask $M_\text{RSP}$ is directly set to all ones. An approximated central view will be reconstructed using their optimal spatial-temporal matches from the six side views (i.e., frame 1,2,3,5,6,7 of the sequence).

The approximation quality measured in PSNR for each data is shown in TABLE \ref{tbl_spReconstruct}. As can be seen, matching based on SP units provides an average \textbf{1.24} \textit{dB} advantage over those based on rectangular patches. Fig. \ref{fig_spReconMiddlebury} shows the visual comparison of reconstructed central views between the two methods. As can be seen, SP provides better approximation especially along occluding boundaries. Reconstruction based on rectangular patches show obvious artifacts at these regions. This experiment validates the advantage of using SP units for content alignment between adjacent frames.

\subsection{Rain Removal Quantitative Evaluations}

To quantitatively evaluate our proposed algorithms, we used a set of 8 videos (different from the training set), and synthesized rain over these videos with varying parameters with \textit{Adobe After Effects}. Each video contains around 200 to 300 frames. All subsequent results shown for each video are the average results for all frames. 

%%%%%%%%%%%%%%%%%%%%%%%Figure%%%%%%%%%%%%%%%%%%%%%%%
\begin{figure}
	\centering
	\includegraphics[width=1\linewidth]{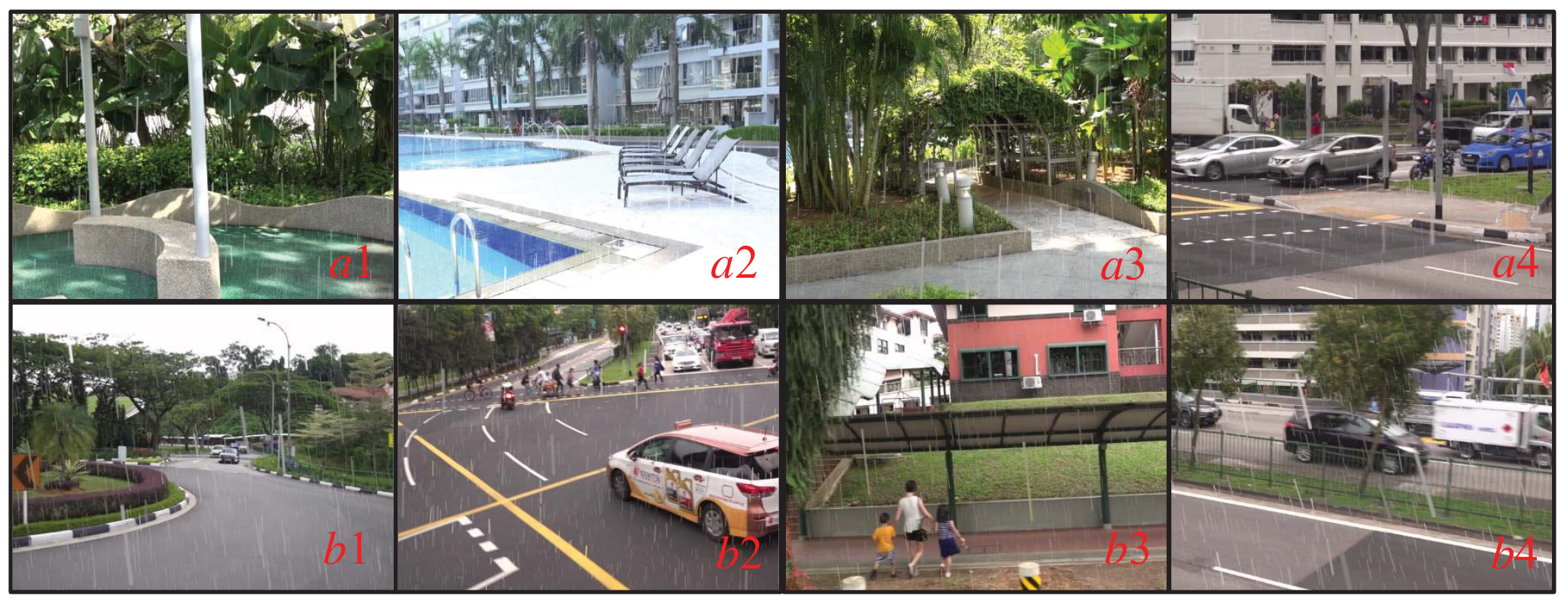}
	\caption{8 testing rainy scenes synthesized with \textit{Adobe After Effects} \cite{adobeAE}. First row (Group \textit{a}) are taken with a panning unstable camera. Second row (Group \textit{b}) are from a fast moving camera (speed range between 20 to 30 \textit{km/h})}
	\vspace{-0.5cm}
	\label{fig_testSet}
\end{figure}
%%%%%%%%%%%%%%%%%%%%%%%Figure%%%%%%%%%%%%%%%%%%%%%%%

To test the algorithm performance in handling cameras with different motion, we divided the 8 testing scenes into two groups: Group \textit{a} consists of scenes taken from a panning and unstable camera with slow movements; Group \textit{b} from a fast moving car-mount camera (with speed range between 20 to 30 \textit{km/h}). Thumbnails and the labeling of each testing scene are shown in Fig. \ref{fig_testSet}.

We evaluated all the methods under the \textit{SPAC} framework, i.e., \textit{SPAC-$\mathcal{F}_1$}, \textit{SPAC-Avg}, \textit{SPAC-RPCA}, and \textit{SPAC-CNN}. Five competing state-of-the-art rain removal methods were chosen for comparison. This includes two image-based method, i.e., 
the discriminative sparse coding (\textit{DSC}) method\footnote{Code available at: \url{http://www.math.nus.edu.sg/~matjh/download/image_deraining/}} \cite{luo2015removing}, and the deep detail network (\textit{DDN}) method\footnote{Code available at: \url{https://xueyangfu.github.io/projects/cvpr2017.html}} \cite{fu2017removing}; and three video-based methods, i.e., rain removal via matrix decomposition (\textit{VMD})\footnote{The authors would like to thank Weihong Ren for sharing the code.} \cite{ren2017video}, via stochastic analysis (\textit{VST})\footnote{Code available at: \url{https://github.com/wwxjtu/RainRemoval_ICCV2017}}\cite{wei2017should}, and via temporal correlation/matrix low-rank completion (\textit{TCL})\footnote{Code available at: \url{http://mcl.korea.ac.kr/deraining/}} \cite{kim2015video}. 
For \textit{VMD}, the number of frames was set to $n=11$ for our test videos as advised by the author.
\textit{VST} assumes the background is static throughout the video sequence, and thus distorts the background severely in our test videos with fast motion.
To improve its performance, we split videos into batches of 10 frames and derain each batch individually.
%Nevertheless, background distortion is still too high to provide comparable derain performance to other methods. Hence, we only include results for PSNR and SSIM.

\subsubsection{Rain Streak Occlusion Precision Recall Rates}\label{sec_pr}

Rain fall introduces edges and textures over the background. Rain removal algorithms are expected to accurately detect and remove these high frequency contents, whilst preserving the edges and textures that belong to the background. 
To evaluate how much of the modifications from the derain algorithms contribute to \textit{\textbf{only removing the rain pixels}}, we calculated the rain streak edge precision-recall (PR) curves.
Absolute difference values were calculated between the derain output against the scene ground truth. Different threshold values were applied to retrieve a set of binary maps, which were next compared against the ground truth rain pixel map to calculate the PR rates. Higher PR value indicates \textit{better rain streak removal quality with less scene distortion}. 

\begin{figure}
	\centering
	\includegraphics[width=1\linewidth]{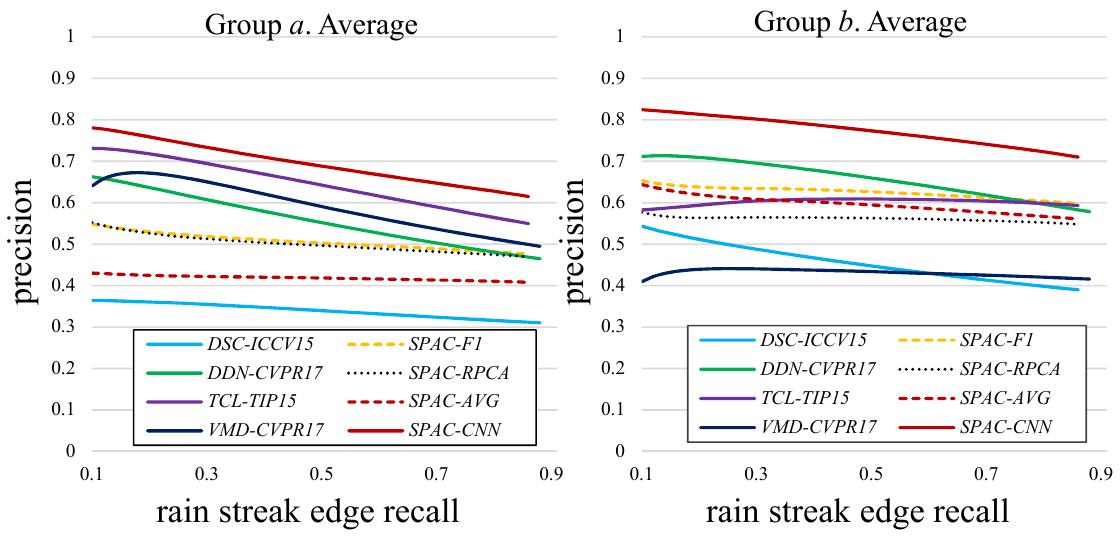}
	\caption{Rain edge pixel detection precision-recall curves for different rain removal methods.}
	\label{fig_prCurves}
\end{figure}

Average PR curves for the two groups of testing scenes by different algorithms are shown in Fig. \ref{fig_prCurves}. 
We can make a few observations.

First, \textit{SPAC-CNN} shows the best performance among all methods in both Group \textit{a} and Group \textit{b}. The performance of \textit{SPAC-CNN} compared with \textit{SPAC-$\mathcal{F}_1$} and \textit{SPAC-Avg} validates the efficiency of the CNN features $\mathcal{F}_2$ and $\mathcal{F}_3$, which correct mis-detected rain streaks (false negative rain detection), and compensate for misalignment blur (false positive rain detection).

Second, \textit{SPAC-RPCA} is competitive with \textit{SPAC-$\mathcal{F}_1$} and outperforms \textit{SPAC-Avg} for Group \textit{a}, but falls behind these methods for Group \textit{b}.
RPCA models the scene background as low-rank, which gives tolerance to very small linear content shifts.
It preserves background details better than \textit{SPAC-Avg} when small camera motion is present.
However, in Group \textit{b} videos, faster camera motion and dynamic scenes reduces the similarity of scene content in Tensor $\mathcal{T}_1$; this violates the low rank assumption required for RPCA, causing \textit{SPAC-RPCA} to perform worse.

Third, for non-SPAC methods, video-based derain methods (\textit{TCL} and \textit{VMD}) perform better than image-based methods (\textit{DSC} and \textit{DDN}) for scenes in Group \textit{a}. 
With slow camera motion, temporal correspondence can be more accurately established, which provides an advantage to video-based methods. 
However, with fast camera motion, temporal correspondence becomes less accurate, and the performance of video-based methods deteriorate relative to image-based methods.

Fourth, the SPAC methods are more robust to fast camera motion than other video-based rain removal methods.
In particular, \textit{SPAC-RPCA} is more robust than \textit{VMD} and \textit{VST}, although all three methods model the background as a low rank matrix.
%Similarly to \textit{VMD} and \textit{VST}, \textit{SPAC-RPCA} models the background as a low rank matrix.
\textit{VST} does not compensate for changes in the background. Hence, as seen in TABLE \ref{tbl_psnr}, it performs poorly in our test videos.
\textit{VMD} relies on global frame alignment to compensate for camera motion; hence, it is robust towards slow camera motion.
\textit{SPAC-RPCA} aligns scene content at a superpixel granularity, which provides the best compensation for camera motion.

%%%%%%%%%%%%%%%%%%%%%%%Figure%%%%%%%%%%%%%%%%%%%%%%%

\subsubsection{Scene Restoration PSNR/SSIM}

We calculated the PSNR, and Structural Similarity (SSIM) \cite{wang2004image} of restored scenes after rain removal between different state-of-the-art methods against the ground truth, and the results are shown in TABLE \ref{tbl_psnr}. % The F-measure for rain streak edge PR curves are also listed for each data. 

Under the \textit{SPAC} framework, \textit{SPAC-RPCA} shows better performance than \textit{SPAC-$\mathcal{F}_1$} and \textit{SPAC-Avg} for Group \textit{a}; however \textit{SPAC-$\mathcal{F}_1$} shows better performance for Group \textit{b}. For both Groups \textit{a} and \textit{b}, \textit{SPAC-CNN} is consistently 5 \textit{dB} higher than \textit{SPAC-Avg}, and  2 \textit{dB} higher than \textit{SPAC-$\mathcal{F}_1$}. SSIM of \textit{SPAC-CNN} is also consistently highest among all \textit{SPAC} methods. This further validates the efficiency of the CNN detail compensation network.

Video based methods (\textit{TCL} and \textit{SPAC-CNN}) outperform image-based methods for Group \textit{a} data (around 2dB and 3dB higher respectively than \textit{DDN}, 3dB and 5dB higher respectively than \textit{DSC}). For Group \textit{b}, image-based methods outperform \textit{TCL} and \textit{VMD}. However, \textit{SPAC-CNN} still holds a 3dB advantage over \textit{DDN}, 4dB over \textit{DSC}.

%%%%%%%%%%%%%%%TABLE%%%%%%%%%%%%%%%%
{\renewcommand{\arraystretch}{1.3}
	\begin{table}[t]
		\begin{center}
			\small\centering\setlength\tabcolsep{1pt}
			\caption{Derain PSNR (\textit{dB}) with different features absent.}
			\label{tbl_evalFeatures}
			\begin{tabular}{|>{\centering\arraybackslash}m{1cm}|>{\centering\arraybackslash}m{1.5cm}|>{\centering\arraybackslash}m{1.5cm}|>{\centering\arraybackslash}m{1.5cm}|>{\centering\arraybackslash}m{1.8cm}|}
				\hline
				\textit{\parbox{0.8cm}{Data \\ Group}} &\parbox{1.3cm}{~~$\hat{\mathcal{F}}_2$+$\hat{\mathcal{F}}_3$ \\ (w/o $\hat{\mathcal{F}}_1$)} &\parbox{1.3cm}{~~$\hat{\mathcal{F}}_1$+$\hat{\mathcal{F}}_3$ \\ (w/o $\hat{\mathcal{F}}_2$)} &\parbox{1.3cm}{~~$\hat{\mathcal{F}}_1$+$\hat{\mathcal{F}}_2$ \\ (w/o $\hat{\mathcal{F}}_3$)} &\parbox{1.5cm}{$\hat{\mathcal{F}}_1$+$\hat{\mathcal{F}}_2$+$\hat{\mathcal{F}}_3$} \\ \hline \hline 
				\textit{a} &27.55 &30.51 &30.22 &\textbf{31.11}\\\hline
				\textit{b} &28.32 &31.86 &32.59 &\textbf{33.19}\\\hline			
			\end{tabular}
		\end{center}
	\end{table}
}
%%%%%%%%%%%%%%%TABLE%%%%%%%%%%%%%%%%

%%%%%%%%%%%%%%%TABLE%%%%%%%%%%%%%%%%
{\renewcommand{\arraystretch}{1.3}
	\begin{table*}[ht]
		\small
		\centering\setlength\tabcolsep{1.8pt} % default value: 6pt
		\begin{center}		
			\caption{Rain removal performance comparison between different methods in terms of scene reconstruction PSNR/SSIM. The best performance for each video sequence have been highlighted in underlined bold letters, and the second best have been highlighted with underlines.} 
			\label{tbl_psnr2}
			\begin{tabular}{|>{\centering\arraybackslash}m{1cm}|>{\centering\arraybackslash}m{0.5cm}|>{\centering\arraybackslash}m{1.48cm}|>{\centering\arraybackslash}m{1.48cm}|>{\centering\arraybackslash}m{1.48cm}|>{\centering\arraybackslash}m{1.48cm}|>{\centering\arraybackslash}m{1.48cm}|>{\centering\arraybackslash}m{1.48cm}||>{\centering\arraybackslash}m{1.48cm}|>{\centering\arraybackslash}m{1.48cm}|>{\centering\arraybackslash}m{1.48cm}|>{\centering\arraybackslash}m{1.48cm}|}
				\hline
				\multirow{3}{*}{\parbox{1cm}{\textit{Camera\\Motion}}} &\multirow{3}{*}{\parbox{0.5cm}{\textit{Clip No.}}}　&\multirow{2}{*}{Rain} &\textit{DSC}~\cite{luo2015removing} &\textit{DDN}~\cite{fu2017removing} &\textit{VMD}~\cite{ren2017video} &\textit{VST}~\cite{wei2017should} &\textit{TCL}~\cite{kim2015video} &\multicolumn{4}{c|}{\textit{SPAC}}\\\cline{9-12} 
				& & &\textit{ICCV15'} &\textit{CVPR17'} &\textit{CVPR17'} &\textit{ICCV17'} &\textit{TIP15'} &\textit{$\mathcal{F}_1$}&\textit{RPCA}&\textit{Avg}&\textit{CNN}\\
				\cline{3-12}
				&　&--&Image &Image &Video &Video &Video &Video &Video &Video &Video  \\
				\hline\hline
				\multirow{5}{*}{\parbox{0.8cm}{\textit{~~slow\\camera\\motion}}} 
				&\textit{a}1 &28.46/0.94 &25.61/0.93 &28.02/0.95 &26.96/0.92 &26.14/0.94 &\underline{\textbf{\textit{29.87}}}/\underline{\textit{0.96}} &27.99/0.95 &29.22/0.96 &24.78/0.87 &\underline{\textit{29.78}}/\underline{\textbf{\textit{0.97}}} \\ 
				&\textit{a}2 &28.09/\underline{\textit{0.95}} &27.11/\underline{\textit{0.95}} &27.38/\underline{\textit{0.95}} &24.80/0.93 &24.03/0.83 &29.01/\underline{\textbf{\textit{0.96}}} &29.27/\underline{\textit{0.95}} &\underline{\textit{29.95}}/\underline{\textbf{\textit{0.96}}} &26.34/0.89 &\underline{\textbf{\textit{30.09}}}/\underline{\textbf{\textit{0.96}}} \\
				&\textit{a}3 &27.84/0.93 &25.08/0.92 &27.41/0.94 &26.45/0.90 &20.50/0.70 &28.82/\underline{\textit{0.95}} &28.01/0.94 &\underline{\textit{29.15}}/\underline{\textit{0.95}} &24.72/0.85 &\underline{\textbf{\textit{29.75}}}/\underline{\textbf{\textit{0.96}}} \\
				&\textit{a}4 &31.48/0.95 &28.82/0.95 &32.47/\underline{\textit{0.97}} &29.55/0.94 &33.41/0.96 &\underline{\textit{34.12}}/\underline{\textbf{\textit{0.98}}} &33.13/\underline{\textit{0.97}} &32.69/\underline{\textit{0.97}} &29.90/0.93 &\underline{\textbf{\textit{34.82}}}/\underline{\textbf{\textit{0.98}}} \\\cline{2-12}
				&$\bar{\textbf{\textit{a}}}$ &28.97/0.94 &26.66/0.94 &28.82/0.95 &26.94/0.92 &26.02/0.86 &\underline{\textit{30.46}}/\textit{\underline{0.96}} &28.42/0.95 &29.44/0.96 &26.44/0.89 &\underline{\textbf{\textit{31.11}}}/\underline{\textbf{\textit{0.97}}} \\ \hline\hline
				\multirow{5}{*}{\parbox{0.8cm}{\textit{camera\\speed\\20-30\\km/h}}} 
				&\textit{b}1 &28.72/0.92 &28.78/0.92 &29.48/\underline{\textbf{\textit{0.96}}} &24.09/0.84 &22.25/0.76 &28.07/0.94 &\textit{\underline{29.85}}/\textit{\underline{0.95}} &29.29/0.94 &26.35/0.89  &\underline{\textbf{\textit{31.19}}}/\underline{\textbf{\textit{0.96}}} \\
				&\textit{b}2 &29.49/0.90 &29.58/0.92 &30.23/0.95 &25.81/0.89 &25.13/0.79 &\textit{\underline{32.41}}/\textit{\underline{0.97}} &31.26/0.96 &31.52/0.96 &28.83/0.93  &\underline{\textbf{\textit{34.05}}}/\underline{\textbf{\textit{0.98}}} \\
				&\textit{b}3 &31.04/0.95 &29.55/0.95 &31.39/\textit{\underline{0.97}} &26.12/0.90 &22.08/0.84 &28.29/0.94 &31.50/0.96 &\textit{\underline{31.71}}/\textit{\underline{0.97}} &29.55/0.94  &\underline{\textbf{\textit{33.73}}}/\underline{\textbf{\textit{0.98}}} \\
				&\textit{b}4 &27.99/0.92 &29.10/0.93 &29.83/\textit{\underline{0.96}} &25.90/0.88 &25.63/0.80 &30.38/0.95 &\textit{\underline{31.92}}/0.95 &31.25/\textit{\underline{0.96}} &28.85/0.92 &\underline{\textbf{\textit{33.79}}}/\underline{\textbf{\textit{0.97}}} \\\cline{2-12}
				&$\bar{\textbf{\textit{b}}}$&29.31/0.92 &29.25/0.93 &30.23/0.96 &25.48/0.88 &23.77/0.80 &29.79/0.95 &\textit{\underline{31.13}}/\textit{\underline{0.96}} &30.94/\textit{\underline{0.96}} &28.40/0.92 &\underline{\textbf{\textit{33.19}}}/\underline{\textbf{\textit{0.97}}} \\\hline 
			\end{tabular}\label{tbl_psnr}
		\end{center}
	\end{table*}
}
%%%%%%%%%%%%%%%TABLE%%%%%%%%%%%%%%%%

\subsubsection{Ablation Studies on CNN features}

We evaluated the significance of different input features to the final derain PSNR over the two groups of testing data. Three baseline CNNs with different combinations of features as input were independently trained for this evaluation. As can be seen from the results in TABLE \ref{tbl_evalFeatures}, a combination of all three features $\hat{\mathcal{F}}_1+\hat{\mathcal{F}}_2+\hat{\mathcal{F}}_3$ provides the highest PSNR. $\mathcal{F}_1$ proves to be the most important feature. Visual inspection on the derain output show both $\hat{\mathcal{F}}_2$+$\hat{\mathcal{F}}_3$ and $\hat{\mathcal{F}}_1$+$\hat{\mathcal{F}}_3$ leave significant amount of un-removed rain. Comparing the last two columns, it can be seen that  $\hat{\mathcal{F}}_3$ works more efficiently for Group \textit{a} than \textit{b} (performance improvement of 0.9 dB and 0.6 dB, respectively), which is understandable since the high frequency features are better aligned for slow cameras, which leads to more accurate detail compensation.

\subsection{Rain Removal Visual Comparisons}

\subsubsection{Visual Comparisons among \textit{SPAC} Methods}

Fig. \ref{fig_visualComp_SPAC} gives visual comparisons for the derain outputs from methods under the proposed \textit{SPAC} framework. Several observations can be made: 

%%%%%%%%%%%%%%%%%%%%%%%Figure%%%%%%%%%%%%%%%%%%%%%%%
\begin{figure*}[!t]
	\centerline{\subfloat{\includegraphics[width=0.96\linewidth]{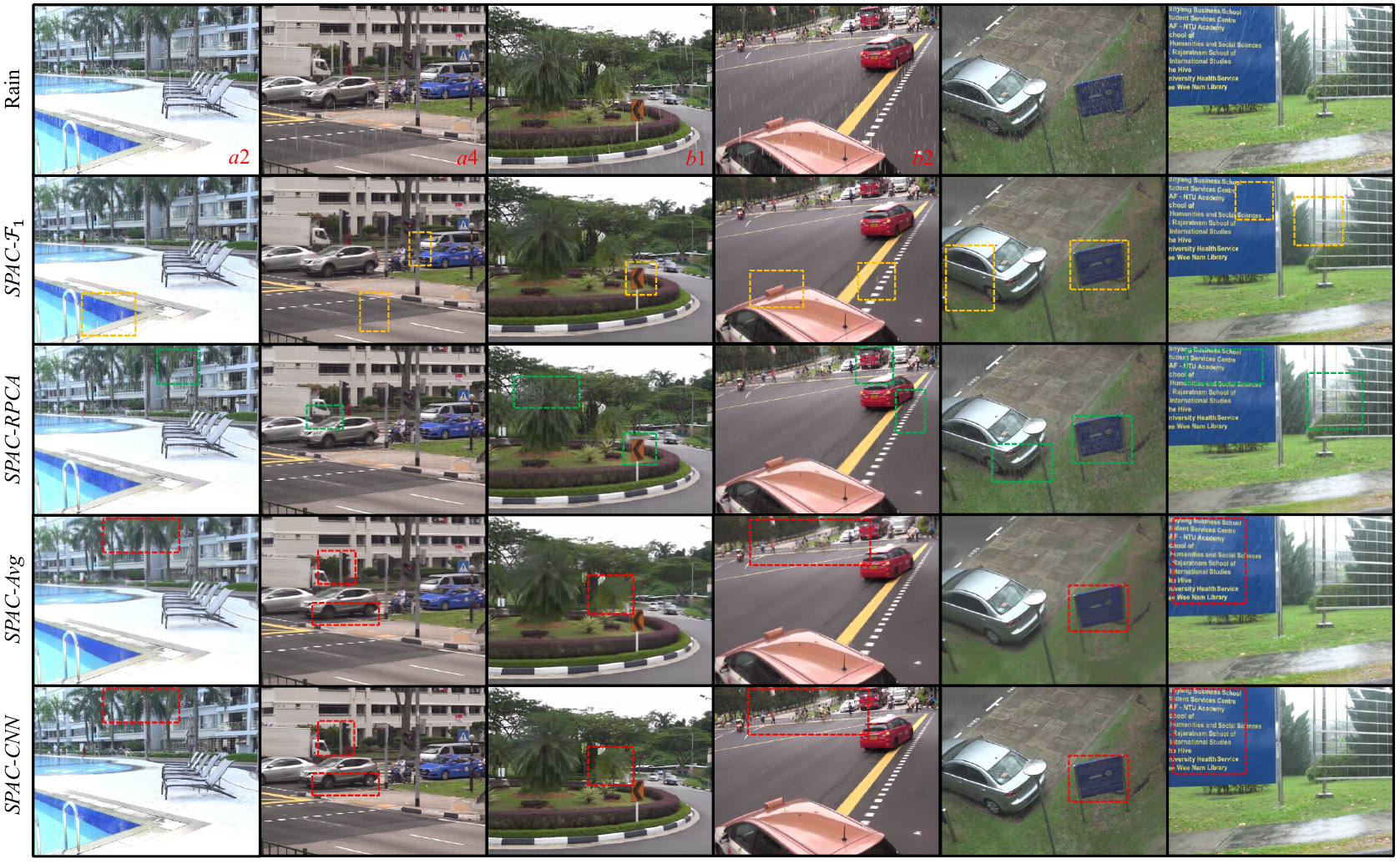}}}
	\caption{Visual comparison for different rain removal methods under the \textit{SPAC} framework for synthetic (column 1 to 4) and real world -rain (column 5 and 6). Electronic zoom-in is advised for better detail comparison.}
	\label{fig_visualComp_SPAC}
\end{figure*}
%%%%%%%%%%%%%%%%%%%%%%%Figure%%%%%%%%%%%%%%%%%%%%%%%

\begin{enumerate}
\item \textit{SPAC-$\mathcal{F}_1$} shows smaller background distortions as compared with \textit{SPAC-RPCA} and \textit{SPAC-Avg}. However, since its output relies on the correctness of the rain mask, remnants of un-detected (false negative) rain streaks can often be found; for the falsely detected rain region (false positive), background distortion (blur) can be observed. These errors have been highlighted in yellow boxes in Fig. \ref{fig_visualComp_SPAC}.
\item The overall quality of rain removal by \textit{SPAC-RPCA} is good. The background is well preserved. There appear to be more, although less opaque, remnant rain streaks compared with \textit{SPAC-$\mathcal{F}_1$}, as highlighted in green boxes. However, it gets worse when rain streaks are dense and opaque (last column).
\item \textit{SPAC-Avg} is able to clear the rain streaks much cleaner than both \textit{SPAC-$\mathcal{F}_1$} and \textit{SPAC-RPCA}, however it seriously blurs the background details.
\item \textit{SPAC-CNN} provides the best visual performance among all \textit{SPAC} methods. The rain is removed most cleanly, and the details are restored truthfully. The enhanced details have been highlighted in red boxes for comparison with \textit{SPAC-Avg}.
\end{enumerate}

\subsubsection{Visual Comparisons with Other Recent Methods}

\begin{figure*}[!t]
	\centerline{\subfloat{\includegraphics[width=0.96\linewidth]{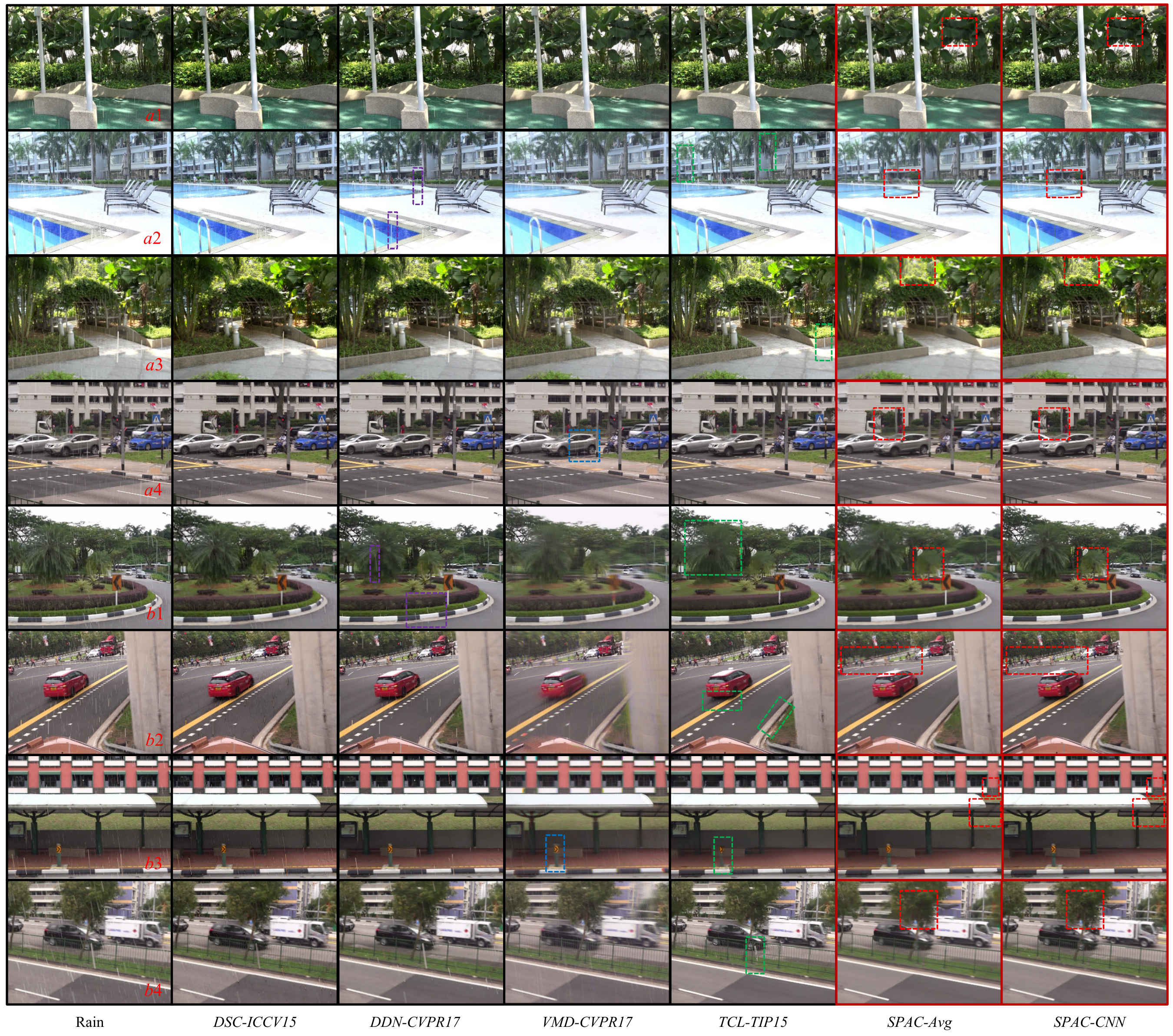}}}
	\caption{Visual comparison between different rain removal methods on the synthetic rain testing dataset. Electronic zoom-in is advised for better detail comparison.}
	\vspace{-0.2cm}
	\label{fig_visualComp_testSet}
\end{figure*}

We carried out visual comparison to compare the derain performance of recent rain removal algorithms. Fig \ref{fig_visualComp_testSet} shows the derain output from different competing methods for all synthetic testing data in Group \textit{a} and \textit{b}. 
%Two consecutive frames are shown for \textit{b}.1 and \textit{b}.4 to demonstrate the camera motion. 

It is observed that rain can be much better removed by video-based methods. Image-based derain methods, i.e., \textit{DSC} \cite{luo2015removing} and \textit{DDN} \cite{fu2017removing}, can only handle light and transparent rain occlusions. For those opaque rain streaks that cover a large area, they fail unavoidably. Temporal information proves to be critical in truthfully restoring the occluded details. 
Both \textit{VMD} \cite{ren2017video} and \textit{TCL} \cite{kim2015video} remove rain perfectly for Group \textit{a}, however they create obvious blur when the camera motion is fast. The derain effect from \textit{SPAC-CNN} is the most impressive for all methods.

\begin{figure*}[!t]
	\centerline{\subfloat{\includegraphics[width=0.96\linewidth]{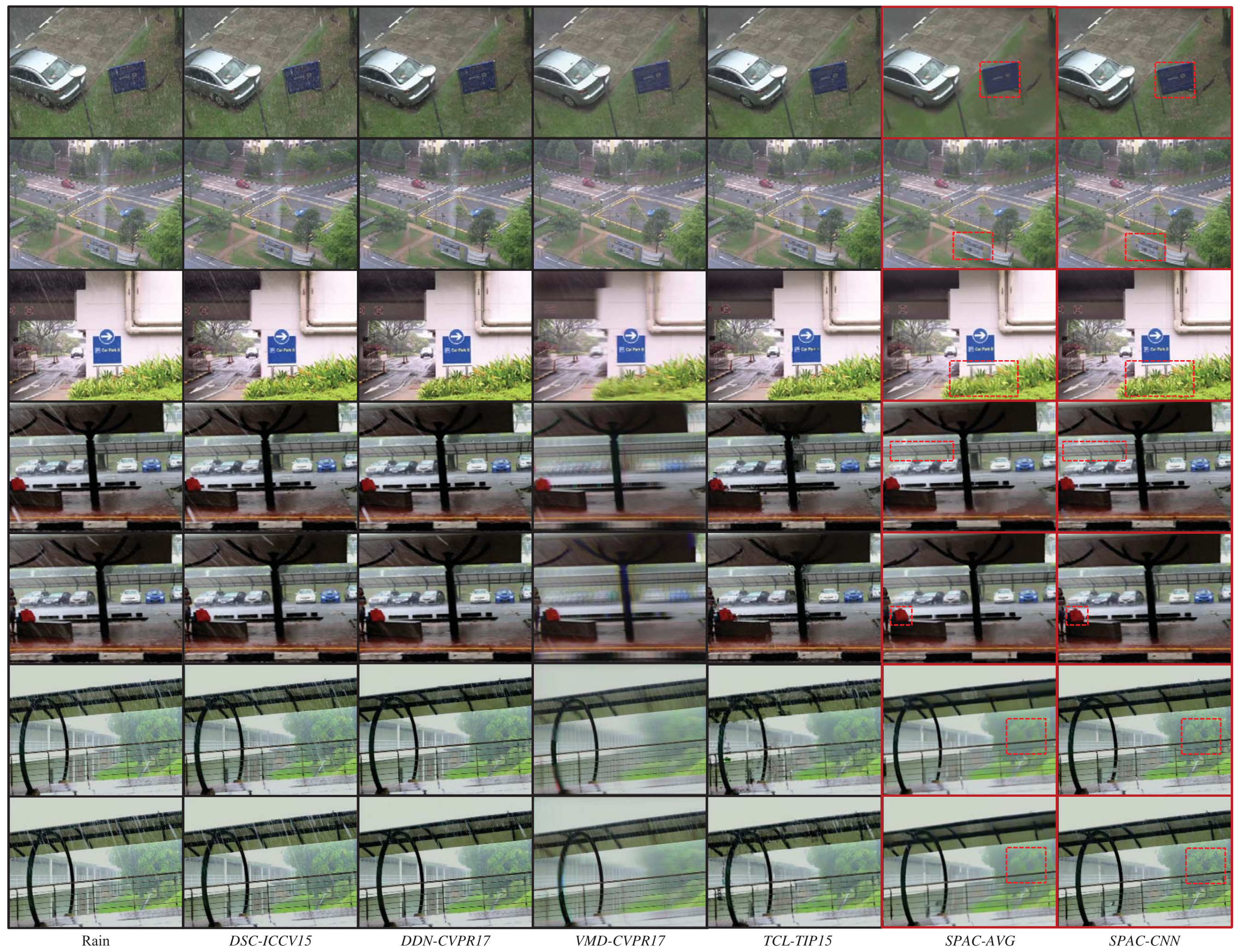}}}
	\caption{Visual comparison between different rain removal methods on real world rain. Electronic zoom-in is advised for better detail comparison.}
	\label{fig_visualComp_realrain}
\end{figure*}

Next, we compare these methods over real rain data. Fig \ref{fig_visualComp_realrain} shows the derain results over five sequences of real world rain videos. Two adjacent frames for the last two videos are shown to demonstrate the fast camera motion. 
It can be seen that, although the \textit{SPAC-CNN} network has been trained over synthetic rain data, it generalizes well to real world rain as well. The advantage of \textit{SPAC-CNN} is very obvious under heavy rain and fast camera motion.

%%%%%%%%%%%%%%%%%%%%%%%Figure%%%%%%%%%%%%%%%%%%%%%%%
%%%%%%%%%%%%%%%%%%%%%%%Figure%%%%%%%%%%%%%%%%%%%%%%%

%%%%%%%%%%%%%%%%%%%%%%%Figure%%%%%%%%%%%%%%%%%%%%%%%
%%%%%%%%%%%%%%%%%%%%%%%Figure%%%%%%%%%%%%%%%%%%%%%%%

\subsubsection{Derain for Veiling \& Accumulated Rain} \label{sec_exp_veiling&Accu}

\begin{figure*}[!t]
	\centerline{\subfloat{\includegraphics[width=0.96\linewidth]{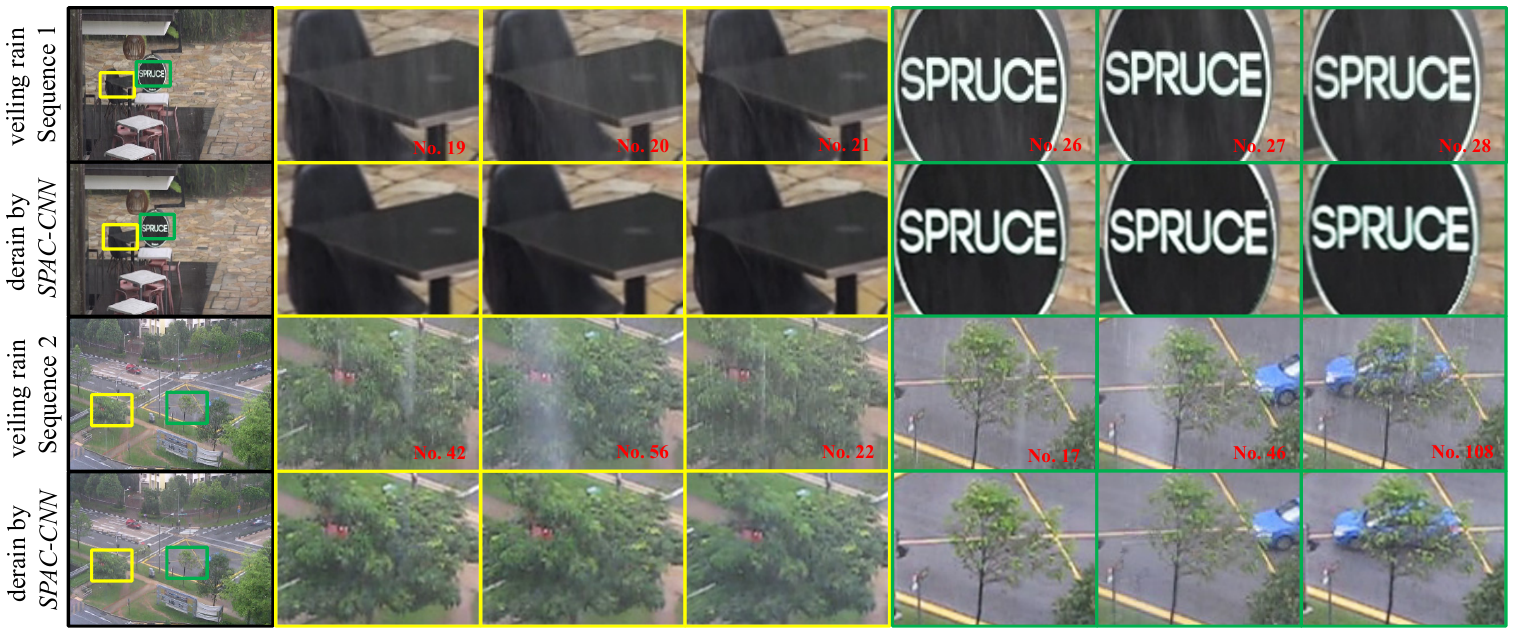}}}
	\caption{Suppression of veiling rain. The first and third rows are zoom-in regions of 6 selected frames from two different video sequences. The second and fourth row are the derain outputs by \textit{SPAC-CNN}.}
	\label{fig_veilingRain}
\end{figure*}

As mentioned in Sec. \ref{sec_T1}, the averaging of $\mathcal{T}_1$ tensor slices is able to suppress both occluding rain streaks, and the out-of-focus rain drops near the camera (veiling rain). Fig. \ref{fig_veilingRain} shows two sequences of videos with veiling rain. The semi-transparent regional intensity fluctuations/distortions may not be apparent when observed in one frame; however, they are clearly visible and distracting in a video sequence. %The affected regions usually appear in a large area, which causes huge problems for both correct rain mask detection, and content alignment. 
Fig. \ref{fig_veilingRain} shows zoom-in regions of 6 selected frames from the two videos. As can be seen, \textit{SPAC-CNN} is able to suppress the fluctuations, and provide a stable clean background, even when the veiling rain covers a large portion of the frame.

%%%%%%%%%%%%%%%%%%%%%%%Figure%%%%%%%%%%%%%%%%%%%%%%%
%%%%%%%%%%%%%%%%%%%%%%%Figure%%%%%%%%%%%%%%%%%%%%%%%

\begin{figure*}[!t]
	\centerline{\subfloat{\includegraphics[width=0.96\linewidth]{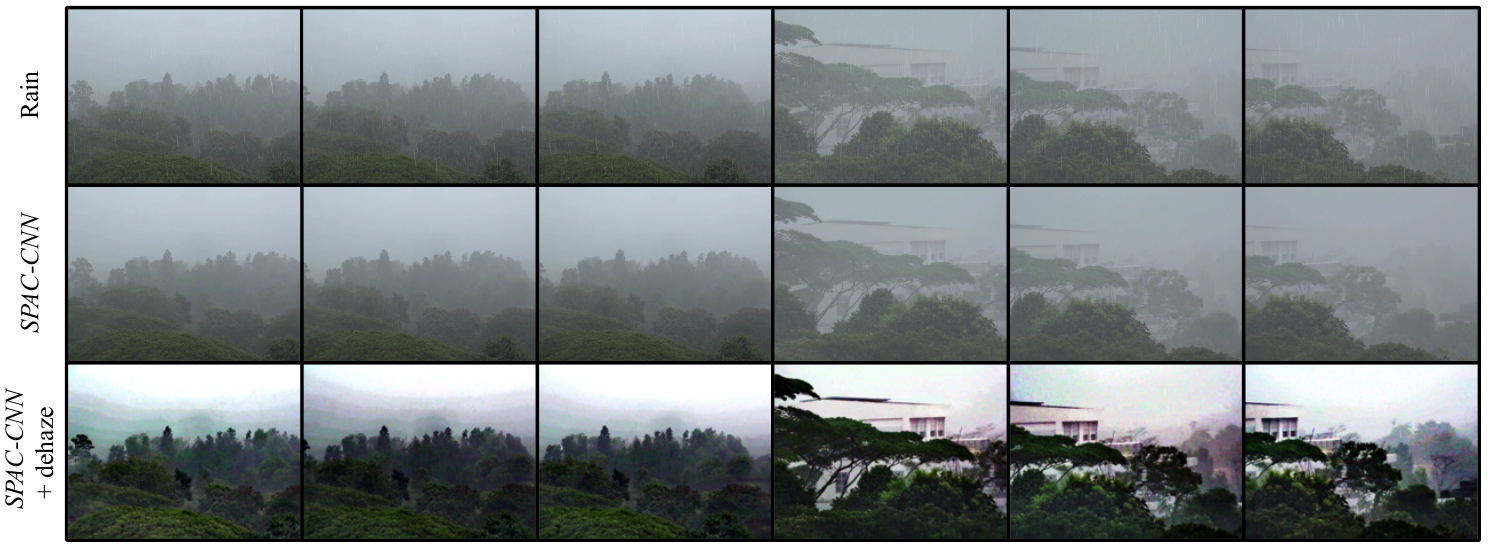}}}
	\caption{Restoration of accumulated rain. The first row are the original rain frames, second row are derain outputs by \textit{SPAC-CNN}, and third row are the non-local \cite{berman2016non} dehazing output applied on the results in the second row.}
	\label{fig_foggyRain}
\end{figure*}

As mentioned in Sec. \ref{sec_introduction}, when the scene distance is large, the rain drops appear as aerosols and accumulate over the distance similar to that of fog and haze. After rain streak removal, dehazing can be applied as an additional step to improve vision contrast. Fig. \ref{fig_foggyRain} shows the dehazing effect using the non-local dehazer \cite{berman2016non} over the output of \textit{SPAC-CNN}. As can be seen, the accumulated rain effect can be effectively removed. The contrast and visibility of distant objects are well restored.

%%%%%%%%%%%%%%%%%%%%%%%Figure%%%%%%%%%%%%%%%%%%%%%%%
%%%%%%%%%%%%%%%%%%%%%%%Figure%%%%%%%%%%%%%%%%%%%%%%%

\subsection{Execution Efficiency}

We compared the average runtime between different methods for deraining a VGA resolution frame. 
We implemented \textit{SPAC-$\mathcal{F}_1$} and \textit{SPAC-Avg} in C++ with GPU acceleration. 
For further acceleration, global frame alignment was applied to bring the majority of scene content into alignment before accelerated template matching, where search was initially limited to a small volume, and later expanded to a larger volume if matches were not found.
\textit{SPAC-RPCA} and \textit{SPAC-CNN}, on the other hand, were implemented in Matlab without the above acceleration methods.
Results are shown in TABLE \ref{tbl_runtime}. As can be seen, \textit{SPAC-$\mathcal{F}_1$} and \textit{SPAC-Avg} are much faster than all other methods. \textit{SPAC-CNN} is much faster than all non-\textit{SPAC} video-based methods and comparable to the fastest image-based method \textit{DDN}. 

%%%%%%%%%%%%%%%TABLE%%%%%%%%%%%%%%%%
{\renewcommand{\arraystretch}{1.3}
	\begin{table}[t]
		\begin{center}
			\small\centering\setlength\tabcolsep{1pt}
			\caption{Execution time (in \textit{sec}) comparison for different methods on deraining a single VGA resolution frame.}
			\vspace{-0.2cm} 
			\label{tbl_runtime}
			\begin{tabular}{|>{\centering\arraybackslash}m{2.7cm}|>{\centering\arraybackslash}m{1.2cm}|>{\centering\arraybackslash}m{0.9cm}||>{\centering\arraybackslash}m{0.35cm}|>{\centering\arraybackslash}m{0.9cm}|>{\centering\arraybackslash}m{1.2cm}|>{\centering\arraybackslash}m{0.9cm}|}
				\hline 
				Method &Platform &Time &\multicolumn{2}{c|}{Method} &Platform &Time\\ \hline\hline
				
				\textit{DSC-ICCV15} \cite{luo2015removing}&Matlab &236.3&\multirow{4}{*}{\rotatebox[origin=c]{90}{SPAC}}& $\mathcal{F}_1$ &C++& 0.2\\\cline{1-3} \cline{5-7} 
				\textit{DDN-CVPR17} \cite{fu2017removing} &Matlab &0.9 & &\textit{RPCA} &Matlab &22.3 \\ \cline{1-3} \cline{5-7} 
				\textit{VMD-CVPR17} \cite{ren2017video} &Matlab &119.0 & &\textit{Avg} &C++& 0.2\\ \cline{1-3}\cline{5-7} 
				\textit{TCL-TIP15} \cite{kim2015video} &Matlab &169.7 & &\textit{CNN} &Matlab &3.3\\ \hline
			\end{tabular}
		\end{center}
		\vspace{-0.6cm}
	\end{table}
}
%%%%%%%%%%%%%%%TABLE%%%%%%%%%%%%%%%%

\section{Conclusion and Discussion}\label{sec_conclusion}
\subsection{Concluding Remarks}
We have proposed a video-based rain removal framework based on SP content alignment and compensation. The proposed framework has been proven to be robust against fast camera motion, and can handle both opaque occluding rain streaks, and semi-transparent veiling rain that exhibits large areas of intensity fluctuations and distortions. 

%Different classical and novel models such as RPCA, and CNN have been applied and compared for their respective advantages in efficiently exploiting the rich spatial-temporal features of the SP alignment outputs.
%The whole system shows its efficiency and robustness over a series of experiments which outperforms state-of-the-art methods significantly.
%The framework of a SP-based alignment module plus a CNN compensation module has potential for various content alignment/registration applications under severe occlusion interference. We plan to try out different possibilities, and further improve its robustness in the future.
% For the derain methods under our proposed  \textit{SPAC} framework, the choice of SP as the basic operation unit is key to its performance. 
% When other decomposition units are used instead (e.g., rectangular), matching accuracy deteriorates, and very obvious averaging blur will be introduced especially at object boundaries.
As the basic processing unit of our proposed algorithms, although the SP template matching can only handle translational motion, alignment errors caused by other types of motion such as rotation, scaling, and non-ridge transforms can be mitigated with global frame alignment before they are buffered (as shown in Fig. \ref{fig_system}) \cite{tan2014dynamic}. Furthermore, these errors can be well modeled as sparse errors by \textit{SPAC-RPCA}, and can be efficiently compensated by \textit{SPAC-CNN}.

\subsection{Future Work}
The good performance of \textit{SPAC} methods rely on the SP alignment accuracy. 
When camera moves even faster, SP search range $n_s$ needs to be enlarged accordingly, which increases computation loads significantly.
We have tested scenarios with camera speed going up to 50 \textit{km/h}, the PSNR becomes lower since corresponding content are less likely be found within the search range. We believe a re-trained CNN with training data from such fast moving camera will help improve the performance. We plan to further investigate rain removal for even faster moving cameras (50 \textit{km/h}+) in the future.

The framework of a SP-based alignment module plus a CNN compensation module has potential for various content alignment/registration applications under severe occlusion interference. We plan to further investigate these possibilities.

%\ifCLASSOPTIONcompsoc
%\else
%\fi

%\section*{Acknowledgment}
%\addcontentsline{toc}{section}{Acknowledgment}
%The research was partially supported by the ST Engineering-NTU Corporate Lab through the NRF corporate lab@university scheme.

\ifCLASSOPTIONcaptionsoff
  \newpage
\fi

%\end{thebibliography}
\bibliographystyle{IEEEtran}
\bibliography{refs}

\begin{thebibliography}{10}
\providecommand{\url}[1]{#1}
\csname url@samestyle\endcsname
\providecommand{\newblock}{\relax}
\providecommand{\bibinfo}[2]{#2}
\providecommand{\BIBentrySTDinterwordspacing}{\spaceskip=0pt\relax}
\providecommand{\BIBentryALTinterwordstretchfactor}{4}
\providecommand{\BIBentryALTinterwordspacing}{\spaceskip=\fontdimen2\font plus
\BIBentryALTinterwordstretchfactor\fontdimen3\font minus
  \fontdimen4\font\relax}
\providecommand{\BIBforeignlanguage}[2]{{%
\expandafter\ifx\csname l@#1\endcsname\relax
\typeout{** WARNING: IEEEtran.bst: No hyphenation pattern has been}%
\typeout{** loaded for the language `#1'. Using the pattern for}%
\typeout{** the default language instead.}%
\else
\language=\csname l@#1\endcsname
\fi
#2}}
\providecommand{\BIBdecl}{\relax}
\BIBdecl

\bibitem{narasimhan2002vision}
S.~G. Narasimhan and S.~K. Nayar, ``Vision and the atmosphere,''
  \emph{International Journal of Computer Vision}, vol.~48, no.~3, pp.
  233--254, 2002.

\bibitem{Garg2007}
K.~Garg and S.~K. Nayar, ``Vision and rain,'' \emph{International Journal of
  Computer Vision}, vol.~75, no.~1, pp. 3--27, 2007.

\bibitem{Garg2005When}
------, ``When does a camera see rain?'' in \emph{IEEE International Conference
  on Computer Vision}, vol.~2, Oct 2005, pp. 1067--1074.

\bibitem{Tripathi2012}
A.~Tripathi and S.~Mukhopadhyay, ``Video post processing: low-latency
  spatiotemporal approach for detection and removal of rain,'' \emph{IET Image
  Processing}, vol.~6, no.~2, 2012.

\bibitem{tan2014dynamic}
C.-H. Tan, J.~Chen, and L.-P. Chau, ``Dynamic scene rain removal for moving
  cameras,'' in \emph{IEEE International Conference on Digital Signal
  Processing}.\hskip 1em plus 0.5em minus 0.4em\relax IEEE, 2014, pp. 372--376.

\bibitem{ren2017video}
W.~Ren, J.~Tian, Z.~Han, A.~Chan, and Y.~Tang, ``Video desnowing and deraining
  based on matrix decomposition,'' in \emph{IEEE Conference on Computer Vision
  and Pattern Recognition}, 2017, pp. 1--4.

\bibitem{Garg2004}
K.~Garg and S.~K. Nayar, ``Detection and removal of rain from videos,'' in
  \emph{IEEE Conference on Computer Vision and Pattern Recognition}, vol.~1,
  2004, pp. 528--535.

\bibitem{zhang2006}
X.~Zhang, H.~Li, Y.~Qi, W.~K. Leow, and T.~K. Ng, ``Rain removal in video by
  combining temporal and chromatic properties,'' in \emph{IEEE International
  Conference on Multimedia and Expo}, July 2006, pp. 461--464.

\bibitem{bossu2011rain}
J.~Bossu, N.~Hauti{\`e}re, and J.-P. Tarel, ``Rain or snow detection in image
  sequences through use of a histogram of orientation of streaks,''
  \emph{International Journal of Computer Vision}, vol.~93, no.~3, pp.
  348--367, Jul 2011.

\bibitem{chen2014rain}
J.~Chen and L.-P. Chau, ``A rain pixel recovery algorithm for videos with
  highly dynamic scenes,'' \emph{IEEE Transactions on Image Processing},
  vol.~23, no.~3, pp. 1097--1104, 2014.

\bibitem{wei2017should}
W.~Wei, L.~Yi, Q.~Xie, Q.~Zhao, D.~Meng, and Z.~Xu, ``Should we encode rain
  streaks in video as deterministic or stochastic?'' in \emph{IEEE
  International Conference on Computer Vision}, Oct 2017, pp. 2535--2544.

\bibitem{luo2015removing}
Y.~Luo, Y.~Xu, and H.~Ji, ``Removing rain from a single image via
  discriminative sparse coding,'' in \emph{IEEE International Conference on
  Computer Vision}, 2015, pp. 3397--3405.

\bibitem{fu2017removing}
X.~Fu, J.~Huang, D.~Z.~Y. Huang, X.~Ding, and J.~Paisley, ``Removing rain from
  single images via a deep detail network,'' in \emph{IEEE Conference on
  Computer Vision and Pattern Recognition}, 2017.

\bibitem{kim2015video}
J.~H. Kim, J.~Y. Sim, and C.~S. Kim, ``Video deraining and desnowing using
  temporal correlation and low-rank matrix completion,'' \emph{IEEE
  Transactions on Image Processing}, vol.~24, no.~9, pp. 2658--2670, Sept 2015.

\bibitem{chen2018robust}
J.~Chen, C.-H. Tan, J.~Hou, L.-P. Chau, and H.~Li, ``Robust video content
  alignment and compensation for rain removal in a {CNN} framework,'' in
  \emph{IEEE Conference on Computer Vision and Pattern Recognition}, 2018.

\bibitem{chen2017patent}
J.~Chen, C.-H. Tan, L.-P. Chau, and H.~Li, ``Rain {R}emoval {B}ased {O}n
  {S}uperpixel {S}patial-{T}emporal {M}atching {A}nd {F}iltering,'' Singapore
  Patent 10\,201\,703\,975P, 2018.

\bibitem{he2009single}
K.~He, J.~Sun, and X.~Tang, ``Single image haze removal using dark channel
  prior,'' in \emph{IEEE Conference on Computer Vision and Pattern
  Recognition}, June 2009, pp. 1956--1963.

\bibitem{fattal2014dehazing}
R.~Fattal, ``Dehazing using color-lines,'' \emph{ACM Transactions on Graphics},
  vol.~34, no.~1, p.~13, 2014.

\bibitem{berman2016non}
D.~Berman, S.~Avidan \emph{et~al.}, ``Non-local image dehazing,'' in \emph{IEEE
  Conference on Computer Vision and Pattern Recognition}, 2016, pp. 1674--1682.

\bibitem{you2016adherent}
S.~You, R.~T. Tan, R.~Kawakami, Y.~Mukaigawa, and K.~Ikeuchi, ``Adherent
  raindrop modeling, detectionand removal in video,'' \emph{IEEE Transactions
  on Pattern Analysis and Machine Intelligence}, vol.~38, no.~9, pp.
  1721--1733, Sept 2016.

\bibitem{eigen2013restoring}
D.~Eigen, D.~Krishnan, and R.~Fergus, ``Restoring an {Image} {Taken} through a
  {Window} {Covered} with {Dirt} or {Rain},'' in \emph{2013 {IEEE}
  {International} {Conference} on {Computer} {Vision}}, Dec. 2013, pp.
  633--640.

\bibitem{kang2012automatic}
L.-W. Kang, C.-W. Lin, and Y.-H. Fu, ``Automatic single-image-based rain
  streaks removal via image decomposition,'' \emph{IEEE Transactions on Image
  Processing}, vol.~21, no.~4, pp. 1742--1755, Apr. 2012.

\bibitem{huang2014self}
D.-A. Huang, L.-W. Kang, Y.-C. Wang, and C.-W. Lin, ``Self-{Learning} {Based}
  {Image} {Decomposition} {With} {Applications} to {Single} {Image}
  {Denoising},'' \emph{IEEE Transactions on Multimedia}, vol.~16, no.~1, pp.
  83--93, Jan. 2014.

\bibitem{sun2014exploiting}
S.-H. Sun, S.-P. Fan, and Y.-C. Wang, ``Exploiting image structural similarity
  for single image rain removal,'' in \emph{2014 {IEEE} {International}
  {Conference} on {Image} {Processing} ({ICIP})}, Oct. 2014, pp. 4482--4486.

\bibitem{chen2014visual}
D.-Y. Chen, C.-C. Chen, and L.-W. Kang, ``Visual depth guided color image rain
  streaks removal using sparse coding,'' \emph{IEEE Transactions on Circuits
  and Systems for Video Technology}, vol.~24, no.~8, pp. 1430--1455, Aug. 2014.

\bibitem{chen2013generalized}
Y.-L. Chen and C.-T. Hsu, ``A {Generalized} {Low}-{Rank} {Appearance} {Model}
  for {Spatio}-temporally {Correlated} {Rain} {Streaks},'' in \emph{2013 {IEEE}
  {International} {Conference} on {Computer} {Vision} ({ICCV})}, Dec. 2013, pp.
  1968--1975.

\bibitem{li2016rain}
Y.~Li, R.~T. Tan, X.~Guo, J.~Lu, and M.~S. Brown, ``Rain streak removal using
  layer priors,'' in \emph{IEEE Conference on Computer Vision and Pattern
  Recognition}, June 2016, pp. 2736--2744.

\bibitem{li2017single}
------, ``Single {Image} {Rain} {Streak} {Decomposition} {Using} {Layer}
  {Priors},'' \emph{IEEE Transactions on Image Processing}, vol.~26, no.~8, pp.
  3874--3885, Aug. 2017.

\bibitem{kim2013single-image}
J.-H. Kim, C.~Lee, J.-Y. Sim, and C.-S. Kim, ``Single-image deraining using an
  adaptive nonlocal means filter,'' in \emph{IEEE International Conference on
  Image Processing}, Sep. 2013, pp. 914--917.

\bibitem{krizhevsky2012imagenet}
A.~Krizhevsky, I.~Sutskever, and G.~E. Hinton, ``Imagenet classification with
  deep convolutional neural networks,'' in \emph{Advances in neural information
  processing systems}, 2012, pp. 1097--1105.

\bibitem{Kim2016accurate}
J.~Kim, J.~K. Lee, and K.~M. Lee, ``Accurate image super-resolution using very
  deep convolutional networks,'' in \emph{IEEE Conference on Computer Vision
  and Pattern Recognition}, June 2016, pp. 1646--1654.

\bibitem{zhang2017beyong}
K.~Zhang, W.~Zuo, Y.~Chen, D.~Meng, and L.~Zhang, ``Beyond a gaussian denoiser:
  Residual learning of deep cnn for image denoising,'' \emph{IEEE Transactions
  on Image Processing}, vol.~26, no.~7, pp. 3142--3155, July 2017.

\bibitem{yang2017deep}
W.~Yang, R.~T. Tan, J.~Feng, J.~Liu, Z.~Guo, and S.~Yan, ``Deep joint rain
  detection and removal from a single image,'' in \emph{IEEE Conference on
  Computer Vision and Pattern Recognition}, 2017, pp. 1357--1366.

\bibitem{jiang2017novel}
T.~X. Jiang, T.~Z. Huang, X.~L. Zhao, L.~J. Deng, and Y.~Wang, ``A {Novel}
  {Tensor}-{Based} {Video} {Rain} {Streaks} {Removal} {Approach} via
  {Utilizing} {Discriminatively} {Intrinsic} {Priors},'' in \emph{2017 {IEEE}
  {Conference} on {Computer} {Vision} and {Pattern} {Recognition} ({CVPR})},
  Jul. 2017, pp. 2818--2827.

\bibitem{Barnum2009Frequency}
P.~C. Barnum, S.~Narasimhan, and T.~Kanade, ``Analysis of rain and snow in
  frequency space,'' \emph{International Journal of Computer Vision}, vol.~86,
  no.~2, p. 256, Jan 2009.

\bibitem{santhaseelan2015utilizing}
V.~Santhaseelan and V.~K. Asari, ``Utilizing local phase information to remove
  rain from video,'' \emph{International Journal of Computer Vision}, vol. 112,
  no.~1, pp. 71--89, 2015.

\bibitem{kim2015mutli}
H.~G. Kim, S.~J. Seo, and B.~C. Song, ``Multi-frame de-raining algorithm using
  a motion-compensated non-local mean filter for rainy video sequences,''
  \emph{Journal of Visual Communication and Image Representation}, vol.~26, pp.
  317--328, Jan. 2015.

\bibitem{kolda2009tensor}
T.~G. Kolda and B.~W. Bader, ``Tensor decompositions and applications,''
  \emph{SIAM review}, vol.~51, no.~3, pp. 455--500, 2009.

\bibitem{Peng2012RASL}
Y.~Peng, A.~Ganesh, J.~Wright, W.~Xu, and Y.~Ma, ``{RASL}: Robust alignment by
  sparse and low-rank decomposition for linearly correlated images,''
  \emph{IEEE Transactions on Pattern Analysis and Machine Intelligence},
  vol.~34, no.~11, pp. 2233--2246, Nov 2012.

\bibitem{bay2008speeded}
H.~Bay, A.~Ess, T.~Tuytelaars, and L.~V. Gool, ``Speeded-{U}p {R}obust
  {F}eatures ({SURF}),'' \emph{Computer Vision and Image Understanding}, vol.
  110, no.~3, pp. 346 -- 359, 2008.

\bibitem{torr2000mlesac}
P.~H. Torr and A.~Zisserman, ``{MLESAC:} a new robust estimator with
  application to estimating image geometry,'' \emph{Computer Vision and Image
  Understanding}, vol.~78, no.~1, pp. 138--156, 2000.

\bibitem{achanta2012slic}
R.~Achanta, A.~Shaji, K.~Smith, A.~Lucchi, P.~Fua, and S.~Susstrunk, ``{SLIC}
  superpixels compared to state-of-the-art superpixel methods,'' \emph{IEEE
  Transactions on Pattern Analysis and Machine Intelligence}, vol.~34, no.~11,
  pp. 2274--2282, 2012.

\bibitem{bergh2012seeds}
M.~Van~den Bergh, X.~Boix, G.~Roig, B.~de~Capitani, and L.~Van~Gool,
  \emph{{SEEDS:} Superpixels Extracted via Energy-Driven Sampling}.\hskip 1em
  plus 0.5em minus 0.4em\relax Springer Berlin Heidelberg, 2012, pp. 13--26.

\bibitem{li2015superpixel}
Z.~Li and J.~Chen, ``Superpixel segmentation using linear spectral
  clustering,'' in \emph{IEEE Conference on Computer Vision and Pattern
  Recognition}, Jun. 2015, pp. 1356--1363.

\bibitem{Candes2011robust}
E.~J. Cand\`{e}s, X.~Li, Y.~Ma, and J.~Wright, ``Robust principal component
  analysis?'' \emph{Journal of the ACM}, vol.~58, no.~3, pp. 11:1--11:37, Jun.
  2011.

\bibitem{yong2017robust}
H.~Yong, D.~Meng, W.~Zuo, and L.~Zhang, ``Robust online matrix factorization
  for dynamic background subtraction,'' \emph{IEEE Transactions on Pattern
  Analysis and Machine Intelligence}, pp. 1--1, 2018.

\bibitem{gao2014block}
Z.~Gao, L.-F. Cheong, and Y.-X. Wang, ``Block-sparse {RPCA} for salient motion
  detection,'' \emph{IEEE Transactions on Pattern Analysis and Machine
  Intelligence}, vol.~36, no.~10, pp. 1975--1987, 2014.

\bibitem{dai2014simple}
Y.~Dai, H.~Li, and M.~He, ``A simple prior-free method for non-rigid
  structure-from-motion factorization,'' \emph{International Journal of
  Computer Vision}, vol. 107, no.~2, pp. 101--122, 2014.

\bibitem{bertsekas1999nonlinear}
D.~P. Bertsekas, \emph{Nonlinear programming}.\hskip 1em plus 0.5em minus
  0.4em\relax Athena scientific Belmont, 1999.

\bibitem{scharstein2003high}
D.~Scharstein and R.~Szeliski, ``High-accuracy stereo depth maps using
  structured light,'' in \emph{IEEE Conference on Computer Vision and Pattern
  Recognition}, vol.~1, June 2003, pp. I--195--I--202.

\bibitem{glorot2010understanding}
X.~Glorot and Y.~Bengio, ``Understanding the difficulty of training deep
  feedforward neural networks,'' in \emph{International Conference on
  Artificial Intelligence and Statistics}, 2010, pp. 249--256.

\bibitem{kingma2014adam}
D.~Kingma and J.~Ba, ``Adam: A method for stochastic optimization,''
  \emph{arXiv preprint arXiv:1412.6980}, 2014.

\bibitem{adobeAE}
\emph{Adobe After Effects Software}, available at
  \url{www.adobe.com/AfterEffects‎}.

\bibitem{Vedaldi2015mat}
A.~Vedaldi and K.~Lenc, ``Matconvnet: Convolutional neural networks for
  matlab,'' in \emph{ACM International Conference on Multimedia}, ser. MM '15,
  2015, pp. 689--692.

\bibitem{wang2004image}
Z.~Wang, A.~C. Bovik, H.~R. Sheikh, and E.~P. Simoncelli, ``Image quality
  assessment: from error visibility to structural similarity,'' \emph{IEEE
  Transactions on Image Processing}, vol.~13, no.~4, pp. 600--612, 2004.

\end{thebibliography}


%\end{IEEEbiography}


%\newpage



% \vfill


\end{document}